\documentclass[runningheads]{llncs}
\usepackage{graphicx}
\usepackage{amsmath,amssymb} 
\usepackage{color}

\begin{document}

\newcommand{\point}{
    \raise0.7ex\hbox{.}
    }


\pagestyle{headings}

\mainmatter

\title{ConvNet-based Depth Estimation, Reflection Separation and Deblurring of Plenoptic Images} 

\titlerunning{Reflection Separation and Deblurring of Plenoptic Images} 

\authorrunning{Paramanand Chandramouli, Mehdi Noroozi and Paolo Favaro} 

\author{Paramanand Chandramouli, Mehdi Noroozi and Paolo Favaro} 
\institute{Department of Computer Science, University of Bern} 

\maketitle

\begin{abstract}
In this paper, we address the problem of reflection removal and deblurring from a single image captured by a plenoptic camera. We develop a two-stage approach to recover the scene depth and high resolution textures of the reflected and transmitted layers. For depth estimation in the presence of reflections, we train a classifier through convolutional neural networks. For recovering high resolution textures, we assume that the scene is composed of planar regions and perform the reconstruction of each layer by using an explicit form of the plenoptic camera point spread function. The proposed framework also recovers the sharp scene texture with different motion blurs applied to each layer. We demonstrate our method on challenging real and synthetic images.
\end{abstract}

\section{Introduction}
When imaging scenes with transparent surfaces, the radiance components present behind and in front of a transparent surface get superimposed. Separating the two layers from a composite image is inherently ill-posed since it involves determining two unknowns from a single equation. Consequently, existing approaches address this problem through additional information obtained by capturing a sequence of images \cite{GuoCVPR14,XueTOG,BrownICCV13}, or by modifying the data acquisition modality \cite{schechner2000separation,agrawal2005removing,kong2014physically}, or by imposing specific priors on layers \cite{LevinSingle04,BrowCVPR14}.

 A light field camera has the ability to obtain spatial as well as angular samples of the light field of a scene from a single image \cite{ng2005light}. With a single light field (LF) image, one can perform depth estimation, digital refocusing or rendering from different view points. This has led to an increased popularity of plenoptic cameras in the recent years \cite{web:lytro,web:ray}. While layer separation from a single image is severely ill-posed in conventional imaging, in light field imaging the problem is made feasible as demonstrated in recent works \cite{wanner2013,arxivWang,JSG15:vmv}. These methods obtain the light field by using a camera array. We propose to use instead microlens array-based plenoptic cameras, because they are more compact and portable. However, both depth estimation and layer separation become quite challenging with a plenoptic camera due to the significantly small baseline  \cite{arxivWang,JSG15:vmv}. Thus, we develop a novel technique to estimate depth and separate the reflected and transmitted radiances from a single plenoptic image. 


Because of merging of intensities from the two layers, the standard multi-view correspondence approach cannot be used for depth estimation. We develop a neural network-based classifier for estimating depth maps. Our classifier can also separate the scene into reflective and non-reflective regions. The depth estimation process has a runtime of only \emph{a few seconds} when run on current GPUs. For recovering scene radiance, we consider that each of the two layers have a constant depth. We relate the observed light field image to a texture volume which consists of radiances from the reflected and transmitted layers through a point spread function (PSF) by taking into account the scene depth values and optics of the plenoptic camera. We solve the inverse problem of reconstructing the texture volume within a regularization framework. While imaging low-light scenes or scenes with moving objects, it is very common for motion blur to occur. If reflections are present in such scenarios, conventional deblurring algorithms fail because they do not model superposition of intensities. However, if such scenes are imaged by a plenoptic camera, our framework can be used to reverse the effect of motion blur. Note that motion deblurring along with layer separation is quite a challenging task because not only the number of unknowns that have to be simultaneously estimated is high but also blind deconvolution is known to be inherently ill-posed.

\begin{figure}[t]
\centering
\begin{minipage}[c]{.31\textwidth}
\centering
\includegraphics[width=\linewidth,height = 125pt]{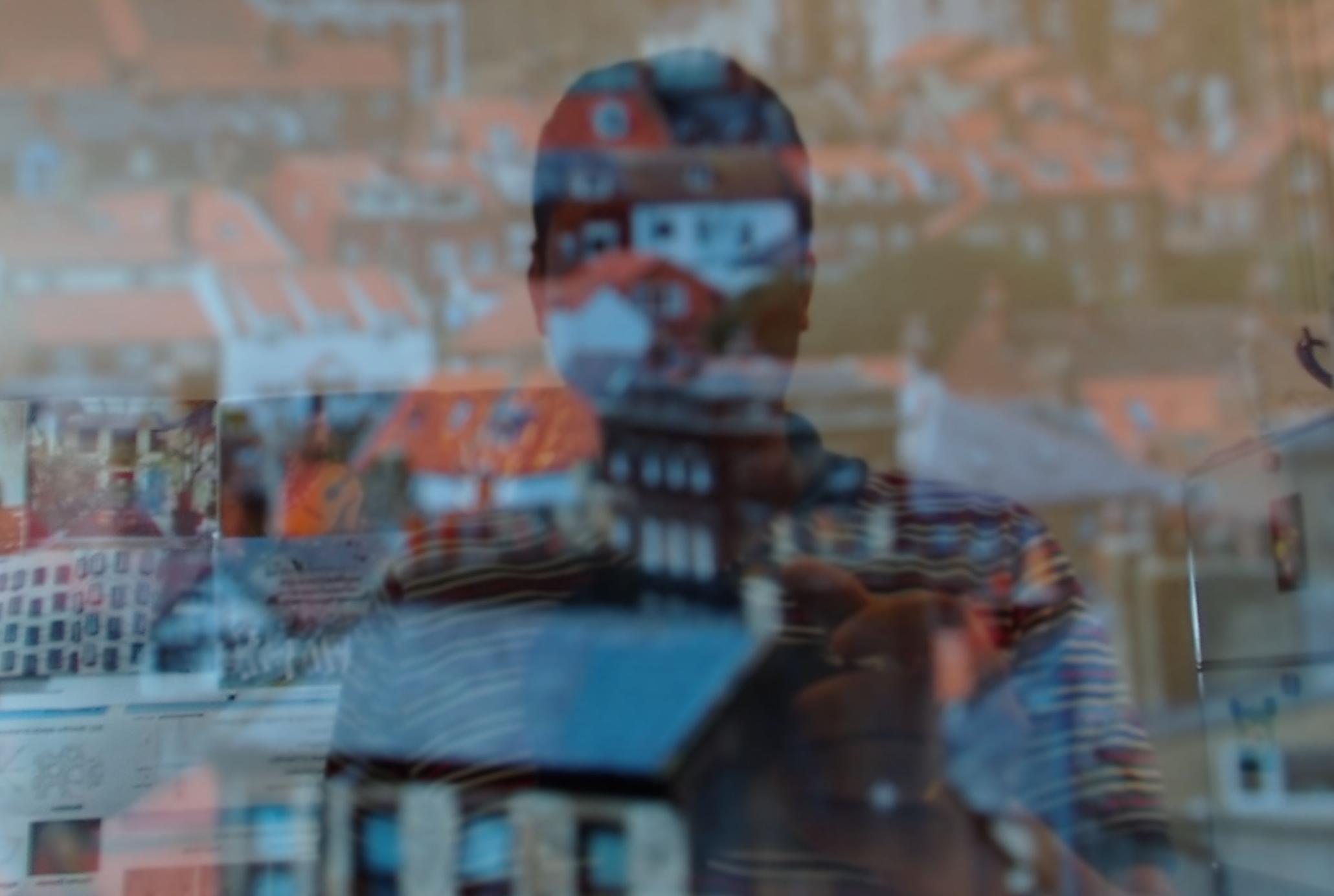} 
\footnotesize{(a)}
\end{minipage}
\begin{minipage}[c]{.31\textwidth}
\centering
\includegraphics[width= \linewidth,height = 125pt]{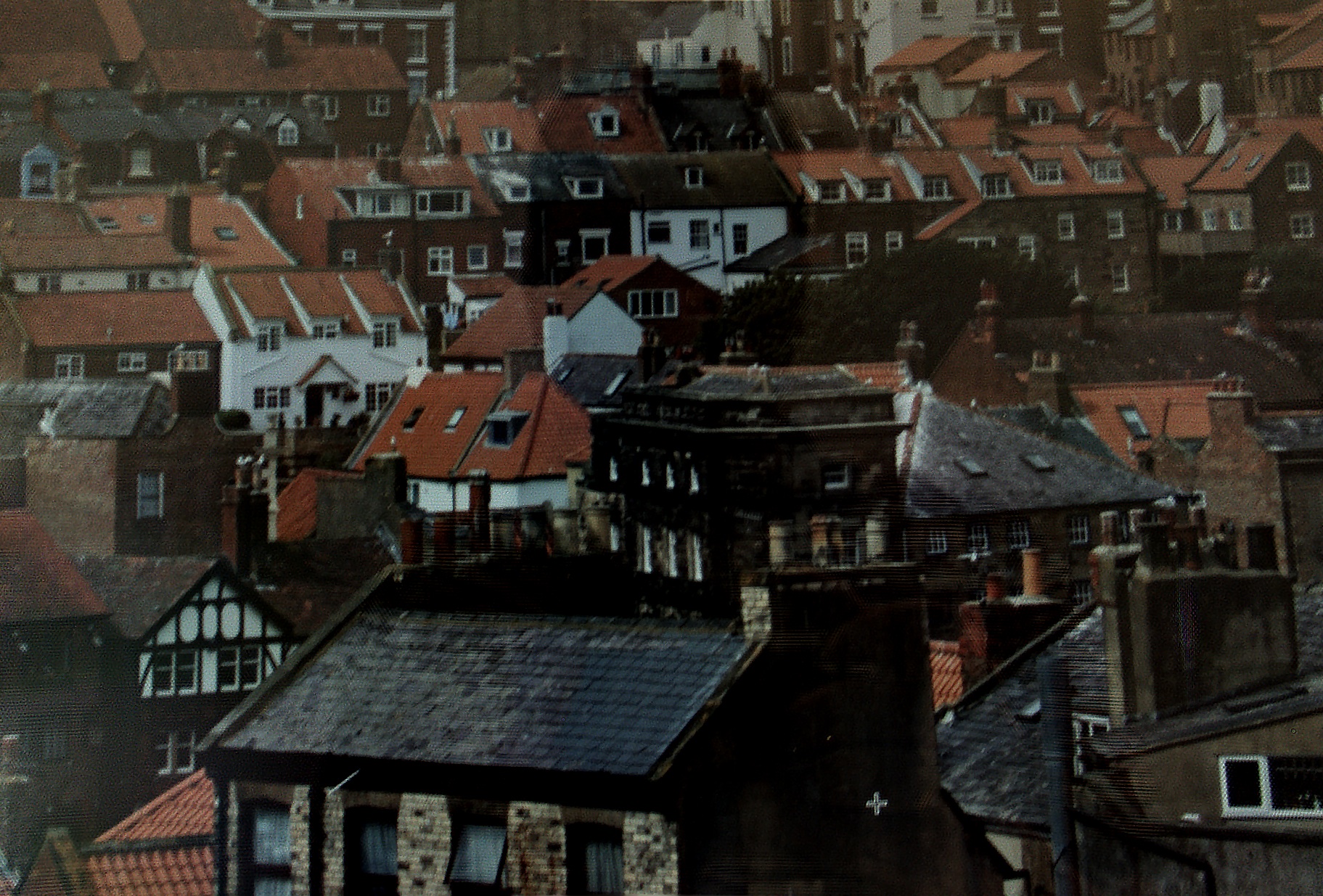}
\footnotesize{(b)}
\end{minipage}
\begin{minipage}[c]{.31\textwidth}
\centering
\includegraphics[width=\linewidth,height = 125pt]{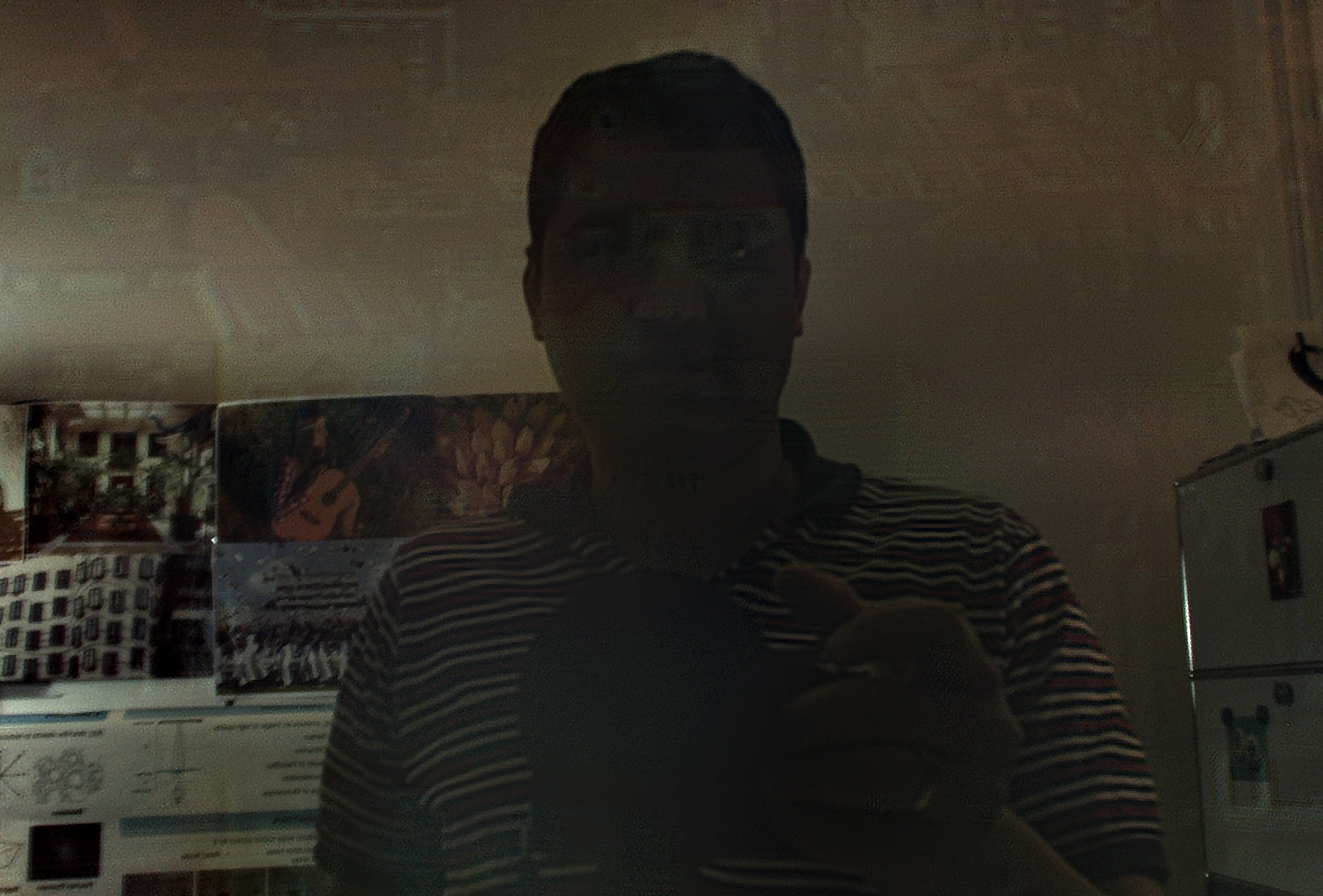} 
\footnotesize{(c)}
\end{minipage}
\hfill
\caption{Layer separation example: (a) Rendered image by Lytro Desktop software. Estimated textures of (b) transmitted and (c) reflected layers.\label{fig:real1}}
\end{figure}

Fig. \ref{fig:real1} shows a real-world example of a scene imaged by a plenoptic camera. One can observe the effect of mixing of intensities of the transmitted and reflected layers in the refocused image generated using Lytro rendering software (Fig. \ref{fig:real1} (a)). Figs. \ref{fig:real1} (b) and (c) show the result of our texture reconstruction algorithm which was preceded by the depth estimation process.

%

\section{Related work}
We briefly review prior works related to reflection separation, plenoptic cameras and motion blur removal.

\textbf{Reflection Separation}
Many techniques make specific assumptions on the scene texture to achieve layer separation from a single image. Levin et al. assume that the number of edges and corners in the scene texture should be as low as possible \cite{LevinSingle04}. In \cite{LevinUser}, a small number of manually labeled gradients corresponding to one layer are taken as input. By using the labeled gradients and a prior derived from the statistics of natural scenes, layer separation is accomplished. Li and Brown develop an algorithm that separates two layers from a single image by imposing different prior distributions on each layer \cite{BrowCVPR14}. Because of the inherent difficulty in the problem, the performance of single image-based techniques is not satisfactory and therefore, additional cues are incorporated \cite{XueTOG}. 

The use of motion parallax observed in a sequence of images has been one of the popular approaches for layer decomposition \cite{GuoCVPR14,XueTOG,BrownICCV13,szeliski2000layer,tsin2006stereo}. As compared to earlier methods the technique of Li and Brown does not enforce restrictions on the scene geometry and camera motion \cite{BrownICCV13}. Based on SIFT-flow they align the images and label the edges as belonging to either of the two layers. The layers are then reconstructed by minimizing an objective function formulated using suitable priors. The recent framework proposed by Xue et al. is applicable for reflection separation as well as occlusion removal without the assumption of a parametric camera motion \cite{XueTOG}. Their method incorporates a robust initialization for the motion field and jointly optimizes for the motion field and the intensities of the layers in a coarse-to-fine manner. In another recent work, Shih et al. propose a layer separation scheme which is based on ghosting effects of the reflected layer \cite{shih2015reflection}.

Wanner and  Goldl{\"u}cke address the problem of estimating the geometry of both the transmitted and reflected surfaces from a 4D light field \cite{wanner2013}. They identify patterns on the epipolar plane image due to reflections and derive local estimates of disparity for both the layers using a second order structure tensor. Wang et al. build an LF array system for capturing light fields \cite{arxivWang}. They estimate an initial disparity map, form an image stack that exhibits low rank property and determine the separation and depth map through an optimization framework based on Robust Principle Component Analysis. The work closest to ours is that of Johannsen et al. \cite{JSG15:vmv}. The technique in \cite{wanner2013} does not lead to good depth estimates on data from real microlens-array based plenoptic cameras. In \cite{JSG15:vmv}, the authors propose improvements to the depth estimation technique in \cite{wanner2013}. They also propose a variational approach for layer separation given the disparities of each layer. The main advantage of our work over this method is that we estimate the radiances of the layers at a much higher resolution. Our method can also be used in the presence of motion blur. Additionally, for real-word data, the technique in \cite{JSG15:vmv} requires user-assisted masks to separate reflecting and Lambertian surfaces. In contrast, our method \emph{automatically} distinguishes between reflective and non-reflective surfaces.


\textbf{Plenoptic cameras}
Although light fields can also be captured from camera arrays, for brevity we focus our discussion on microlens array-based cameras. Ng et al. proposed a portable design for capturing light fields by placing a microlens array between the camera lens and the sensors \cite{ng2005light}. This design also enables post capture refocusing, rendering with alternate viewpoint \cite{ng2005light} and depth estimation \cite{Bishop}. To overcome the limited spatial resolution in plenoptic images, superresolution techniques have been proposed \cite{Bishop,GeorgievSR}. Georgiev et al. \cite{GeorgievSR} directly combine the information from different angular views to obtain a high resolution image. Bishop and Favaro \cite{Bishop} relate the LF image with the scene radiance and depth map through geometric optics. They initially estimate the depth map and subsequently the high resolution scene radiance through deconvolution. In \cite{ppr:Goldluecke13}, along with disparity estimation, an input light field is super-resolved not only in spatial domain but also in angular domain.

In recent years, decoding, calibration and depth estimation techniques have been developed for commercially available cameras. Cho et al. \cite{ppr:calib_tai} develop a method for rectification and decoding of light field data and render the super-resolved texture using a learning-based interpolation method. The calibration techniques in \cite{ppr:calib,ppr:calib_eccv}, develop a model to relate a pixel of the light field image with that of a ray in the scene. A significant number of depth estimation algorithms exist for plenoptic cameras. Most of these methods assume that the scene consists of Lambertian surfaces \cite{ppr:tao,ppr:lf_eccvw,yu2013line,jeon2015accurate,wang2015occlusion}. In a recent work, Tao et al. propose a scheme for depth estimation and specularity removal for both diffuse and specular surfaces \cite{TaoPami2015}. Since view correspondences do not hold for specular surfaces, based on a dichromatic model, they investigate the structure of pixel values from different views in  color space. Based on their analysis, they develop schemes to robustly estimate depth, determine light source color and separate specularity. These techniques cannot be used for reflective surfaces since they do not handle superposition of radiances from two layers.


\textbf{Motion Deblurring}
Recovering the sharp image and motion blur kernel from a given blurry image has been widely studied in the literature \cite{Fergus2006,Levin2011}. Although the blind deconvolution problem is inherently ill-posed, remarkable results have been achieved by incorporating suitable priors on the image and blur kernel \cite{Cho2009,Xu2010,L0,Perrone2014}. While the standard blind deconvolution algorithms consider the motion blur to be uniform across the image, various methods have been proposed to handle blur variations due to camera rotational motion \cite{ppr:whyte,ppr:gupta_eccv,ppr:hirsch_iccv}, depth variations \cite{HuDepth,paramCVPR,Sorel}, and dynamic scenes \cite{Hyun2013}. Although there have been efforts to address the issue of saturated pixels in the blurred observation \cite{whyte2014deblurring}, no deblurring algorithm exists that can handle the merging of image intensities due to transparent surfaces.

Our contributions can be summarized as follows: i) We model the plenoptic image of a scene that contains reflected and transmitted layers by taking into account the effect of camera optics and scene depth. ii) Ours is the first method that uses convolutional neural networks for depth estimation on images from plenoptic cameras. Our classifier can efficiently estimate the depth map of both the layers and works even when there are no reflections. iii) Our PSF-based model inherently constrains the solution space thereby enabling layer separation. iv) We also extend our framework to address the challenging scenario of joint reflection separation and motion deblurring.
\section{Plenoptic image formation of superposed layers}
\label{sec:model}

Initially, let us consider a constant depth scene without any reflections. The LF image of such a scene can be related to the scene radiance through a point spread function (PSF), which characterizes the plenoptic image formation process \cite{Bishop,Broxton:13,Liang:2015}. The PSF is dependent on the depth value and the camera parameters, and encapsulates attributes such as disparity across sub-aperture images and microlens defocus. Mathematically, one can express the light field image $l$ formed at the camera sensors in terms of the scene radiance $f_d$ and a PSF $H_d$ as a matrix vector product. Note that both $l$ and $f_d$ are vectorial representations of a 2D image. The subscript $d$ in the PSF $H_d$ indicates the depth label. The entries of the matrix $H_d$ can be explicitly evaluated from the optical model of the camera \cite{Bishop,Broxton:13}. The PSF $H_d$ is defined by assuming a certain resolution for the scene texture. Defining the scene texture on a finer grid would lead to more columns in the PSF $H_d$. A column of the matrix denotes an LF image that would be formed from a point light source at a location corresponding to the column index. Note that in practice, the light field image and the texture will be of the order of millions of pixels and generating PSF matrices for such sizes is not feasible. However, the PSF has a repetitive structure because the pattern formed by the intersection of the blur circles of the microlens array and the main lens gets repeated. Based on this fact, the matrix vector product $H_d f_d$ can be efficiently implemented through a set of parallel convolutions between the scene texture and a small subset of the elements of the matrix $H_d$ \cite{Broxton:13}.

While imaging scenes consisting of transparent surfaces, radiances from two different layers get superimposed. Let $f_{d_t}$ and $f_{d_r}$ denote the radiances of the transmitted and reflected layers, and $d_t$ and $d_r$ denote the depth values of the transmitted and reflected layers, respectively. Then the LF image $l$ can be expressed as
\begin{align}
{l}=H_{d_t}f_{d_t} + H_{d_r}f_{d_r} = Hf
\label{eqn:lfvec}
\end{align}
where the variable $f$ which is referred to as texture volume is composed of radiances of the two layers, i.e., $f = [f^T_{d_{t}}~ f^T_{d_{r}}]^T$. The matrix $H$ is formed by concatenating the columns of PSF matrices corresponding to depths $d_t$ and $d_r$, i.e., $H = [H_{d_t}~H_{d_r}]$.


According to the imaging model, the plenoptic images corresponding to the transmitted layer lie in the subspace spanned by the columns of $H_{d_t}$, span$\{H_{d_t}\}$, and those of the reflected layer lie in span$\{H_{d_r}\}$. The attributes of an LF image such as extent of blurring and disparity between sub-aperture images vary as the depth changes. We assume that the depth values of the transmitted and the reflected layers are quite different.

\section{Proposed method}
 In real scenarios, the local intensities of an LF image can consist of components from either the transmitted layer or the reflected layer, or both. Due to superposition of the two layers, the standard approach of establishing correspondences across views would not be applicable. To detect the depth, we train a classifier using a convolutional neural network (ConvNet). Subsequently, we solve the inverse problem of estimating the high-resolution textures of the two layers within a regularization framework. In our texture estimation procedure, we consider that the two layers have constant depth. However, our depth estimation technique is applicable even if there are depth variations within each layer. 
\subsection{Depth estimation}
In our experiments, we use the Lytro Illum camera. Consequently some of the details of our depth estimation scheme are specific to that camera. However, our method can be easily adapted to other plenoptic cameras as well. We consider that the scene depth ranges from $20$cm to $2.5$m. The depth range is divided into $15$ levels denoted by the set $\Delta = \{d_1, d_2, \dots, d_N\}$ ($N=15$). The quantization of depth range is finer for smaller depth values and gets coarser as the values increase. For a pair of depths, as their magnitudes increase, their corresponding PSFs become more indistinguishable. Beyond $2.5$m there would be hardly any variations in the PSF.

A patch of an LF image can have intensities from either only one layer (if there is no reflection) or from two layers. We define a set of labels $\Lambda=\{\lambda_1, \lambda_2, \dots \lambda_L\}$ wherein each label denotes either a combination of a pair of depths or individual depths from the set $\Delta$. We assume that the two layers have significantly different depth values and do not include all possible pairs from $\Delta$ in $\Lambda$. Instead, we choose $45$ different labels in the set $\Lambda$.

 We perform depth estimation on patches of 2D raw plenoptic images in which the micolenses are arranged on a regular hexagonal grid and avoid the interpolation effects that occur while converting the 2D image to a 4D representation \cite{ppr:calib}. In an aligned 2D plenoptic image of Lytro Illum camera, the microlens arrangement pattern repeats horizontally after every $16$ pixels and vertically every $28$ pixels. Consequently, we define one unit to consist of $28{\times}16$ pixels. For both training and evaluation we use a patch from an LF image consisting of $10 {\times}10$ units. We convert a patch into a set of views wherein, each view is obtained by sampling along horizontal and vertical directions. We discard those views that correspond to the borders of the microlenses. This rearranged set of views is considered as the input for which a label is to be assigned. Including the three color channels, the dimensions of the input corresponding to a patch of a plenoptic image are $10{\times}10{\times}888$. This input data contains disparity information across the $888$ views similar to a set of sub-aperture images (illustrated in the supplementary material). Our ConvNet is trained using labeled inputs in this format.
\begin{figure}[t]
\begin{center}
\begin{tabular}{cc}
\includegraphics[width=340pt]{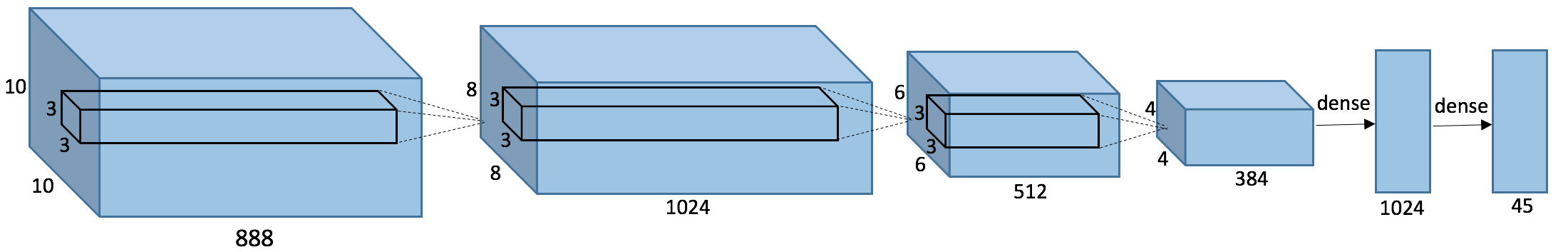}
\end{tabular}
\caption{ConvNet architecture used for depth estimation. Excluding the last layer, all other layers are followed by ReLu (not indicated for the sake of brevity). Since convolutions are performed without padding, and with a stride of one, spatial extent of the data decreases by two after each convolutional layer.  \label{fig:cnn}}
\end{center}
\end{figure}

\subsubsection{Network architecture}
As depicted in the Fig. \ref{fig:cnn} our ConvNet contains five layers. The first three are convolutional, and the fourth and fifth layers are fully connected. No pooling layer is used. We use $3\times3$ filters in the convolutional layers without padding. Our choice of the size of filters leads to a receptive field of $7{\times}7$. This in turn corresponds to an area in the LF image that is sufficiently large enough to capture features.
 
\subsubsection{Training}
For generating the training data, we used real LF images. Using a projector, we displayed a set of natural images on a planar surface. The LF images were captured by placing the camera at distances $d_1, d_2, \dots, d_N$. To obtain the data for a label $\lambda$, if the label corresponded to a combination of two depths, we superimposed the LF images of the two depths, else we directly used patches from the captured images. Our training dataset consisted of about $16{,}000$ patches per label. We implemented stochastic gradient descent with a batch size of $256$ patches. We used the step strategy during the training with 0.01 as the baseline learning rate, and $50{,}000$ as step. The training converged after $170{,}000$ iterations using batch normalization \cite{bn} and took about $10$ hours with one Titan X.  

\subsubsection{Efficient evaluation}
Given an LF image as the input, from the ConvNet, we arrive at a label map which contains the label for each unit ($28{\times}16$ pixels). 
One could follow a straightforward approach of evaluating the label for every patch through the ConvNet. This would involve cropping patches from the LF image through a sliding window scheme (with a shift of one unit), rearranging every patch in the format of dimensions $10{\times}10{\times}888$ and feeding the cropped patches to the network as input. As there is overlap corresponding to one unit amongst these patches, this approach involves redundant time consuming calculations in the convolutional layers and thereby is inefficient. We address this problem by separating convolutional and fully connected layer operations. Instead of cropping patches, we rearrange the entire LF image into a set of views (following a procedure similar to that of a patch) and drop the views corresponding to the borders of the microlenses. This gives us a set of views that are large in size and having dimensions $W \times H \times 888$. Note that each $10\times 10 \times 888$ subregion of this array corresponds to one patch in the original LF image and vice versa. We feed this large array as input to the convolutional layers. The last (third) convolutional layer feature map would be of size $(W-6) \times (H-6) \times 384$ (refer to Fig. \ref{fig:cnn}). To determine the label of a patch in the LF image we find its corresponding $4 \times 4 \times 384$ subregion in the third convolutional layer feature map and feed it to the fully connected layers as input. With this alternate approach, we can calculate the depth map of the full LF image of size of $6{,}048 \times 8{,}640$ pixels in about $3$ seconds with one Titan X.

We convert the labels to two depth maps by assuming that the depth values of one layer is always greater than the other. The non-reflective regions are also automatically indicated by labels that correspond to individual depth entries.
\subsection{Texture reconstruction}
Our texture reconstruction algorithm is restricted to scenes wherein the two layers can be approximated by fronto-parallel planes. We evaluate the median values of the two depth maps to arrive at the depth values $d_t$ and $d_r$. The PSF entries are then evaluated with the knowledge of camera parameters \cite{Bishop}. We define the texture resolution to be one-fourth of the sensor resolution. We formulate a data fidelity cost in terms of the texture volume through the correct PSF. We impose total variation (TV) regularization for each layer separately and arrive at the following objective function
\begin{align}
\min_{f} \left\|l-Hf\right\|^2 +{\nu} \|f_{d_t}\|_{TV} + {\nu}\|f_{d_r}\|_{TV}
\label{eqn:texCost}
\end{align}
where $\|{\cdot}\|_{TV}$ denotes total variation and ${\nu}$ is the regularization parameter. We minimize eq. (\ref{eqn:texCost}) by gradient descent to obtain the texture volume, which consists of textures corresponding to each layer.

In our method, the PSF and TV prior impose constraints on the solution space of the
layer separation problem. As an example, consider that the true texture $\hat{f}_{d_t}$ has an edge at a particular location. The PSF corresponding to depth $d_t$ enforces that the edge gets repeated across $k_{d_t}$ microlenses in the LF image corresponding to depth $d_t$. For the other layer, the number of microlenses, $k_{d_r}$, in which a feature gets repeated would be quite different from $k_{d_t}$, since we consider significant depth differences across layers. Similarly, the other attributes, such as microlens blur and disparity across views, also vary with depth. In our formulation we look for the texture volume that best explains the observed LF image. In the inverse problem of texture volume estimation, the constraints inherently imposed by the PSF avoids those solutions that generate attributes different from that of the observed LF image.
\subsection{Motion blur scenario}
A relative motion between the scene and the camera would lead to a motion blurred light field image. We model the blurry LF image as
\begin{align}
{l}=H_{d_t}f_{d_t} + H_{d_r}f_{d_r} = H_{d_t}M_tu_t + H_{d_r}M_r u_r
\label{eqn:lfm}
\end{align}
where $u_t$ and $u_r$ denote the sharp texture, and $M_t$ and $M_r$ denote the blurring matrices of the transmitted and reflected layers respectively. We consider that the blur is uniform and the matrix vector products ($f_{d_t} = M_t u_t$ and $f_{d_r} = M_r u_r$) denote convolutions.  In this scenario, we can write the objective function in terms of the sharp texture layers as
\begin{align}
\min_{u_t,u_r} \left\|l-H_{d_t}M_tu_t - H_{d_r}M_ru_r\right\|^2 +{\nu} \|u_{t}\|_{TV} + {\nu}\|u_{r}\|_{TV}.
\label{eqn:texCostm}
\end{align}
The objective function in terms of the motion blur kernels is given by
\begin{eqnarray}
\nonumber
\min_{m_t,m_r} \left\|l-H_{d_t}U_tm_t - H_{d_r}U_rm_r\right\|^2 \\
\mbox{subject to }  m_t \succcurlyeq 0, m_r \succcurlyeq 0, \quad \|m_t\|_1 = 1,\|m_r\|_1 = 1
\label{eqn:bCost}
\end{eqnarray}
where $U_t$ and $U_r$ are matrices corresponding to the textures $u_t$ and $u_r$, respectively, and $m_t$ and $m_r$ denote the vectors corresponding to motion blurs kernels of the transmitted and reflected layers, respectively.
We minimize the objective function given in eqs. (\ref{eqn:texCostm}) and (\ref{eqn:bCost}) by following an approach similar to the projected alternating minimization algorithm of \cite{Perrone2014}. In real scenarios, the reflected layer also undergoes the effect of ghosting due to the optical properties of the surface being imaged \cite{shih2015reflection}. The combined effect of ghosting and motion blur can lead to significantly large shifts. Hence deblurring of the reflected layer may not work well in practical situations. However, our model works well even when there is ghosting.


\begin{figure}[t]
\begin{center}
\begin{tabular}{cc}
\begin{tabular}{cc}
\includegraphics[width=72pt,height = 72pt]{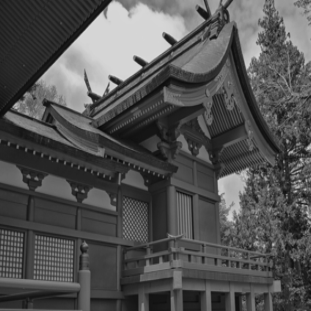}&
\includegraphics[width=72pt,height = 72pt]{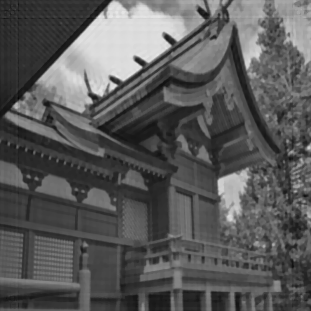}\\
\footnotesize(a)&\footnotesize(c)\\
\includegraphics[width=72pt,height = 72pt]{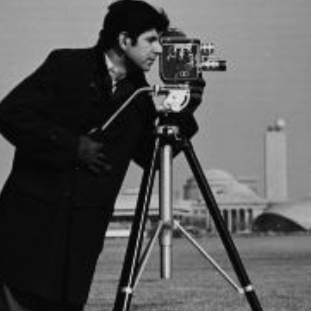}&
\includegraphics[width=72pt,height = 72pt]{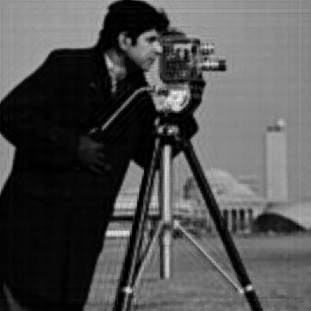}\\
\footnotesize(b)&\footnotesize(d)
\end{tabular}
\begin{tabular}{c}
\includegraphics[width=167pt,height = 160pt]{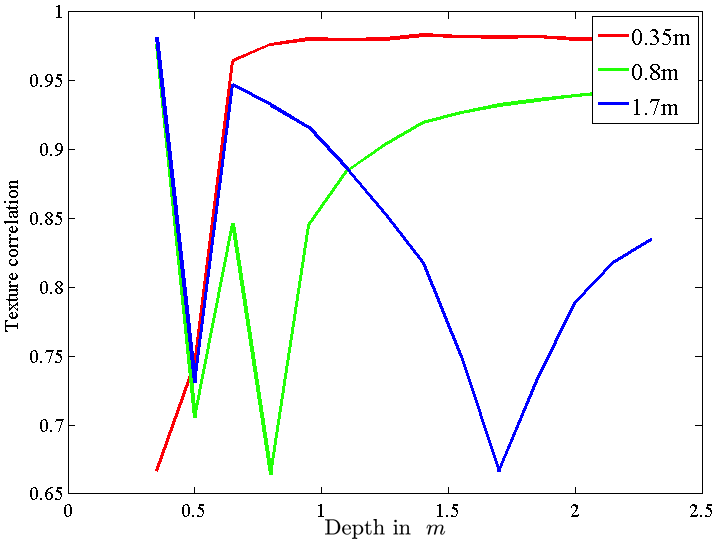}\\
\footnotesize(e)\\
\end{tabular}
\end{tabular}
\caption{Synthetic experiment: (a) and (b) true textures. (c) and (d) recovered textures. (e) Mean correlation between true and recovered texture at different depths (legend indicates the depth value of the transmitted layer). \label{fig:corr}}
\end{center}
\end{figure}

\begin{figure*}[t]
\begin{center}
\begin{tabular}{cccc}
\includegraphics[width=86pt,height = 70pt]{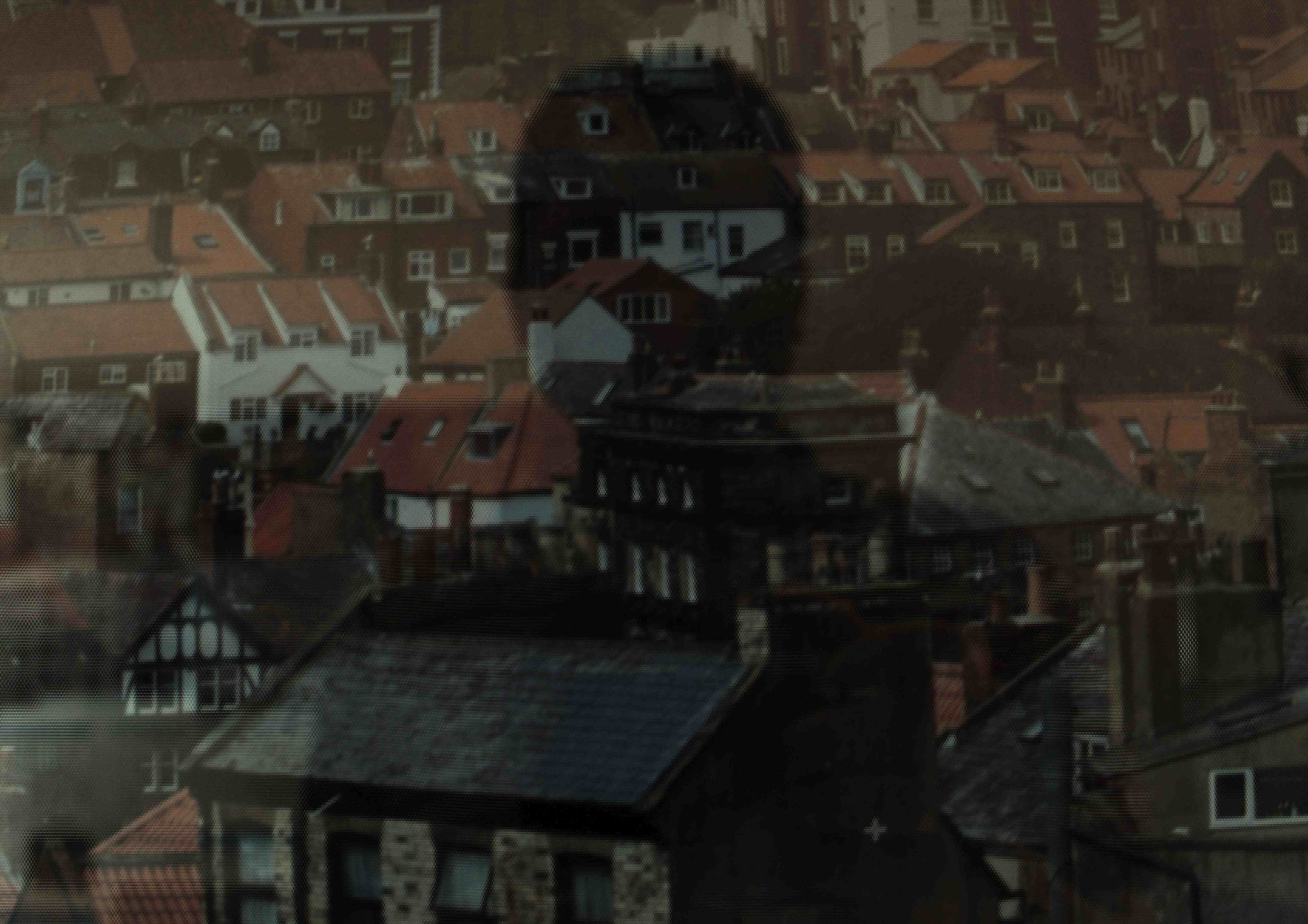} &
\includegraphics[width=86pt,height = 70pt]{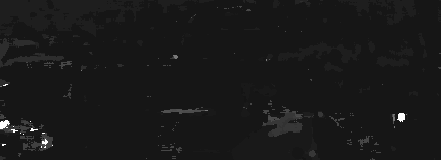}&
\includegraphics[width=86pt,height = 70pt]{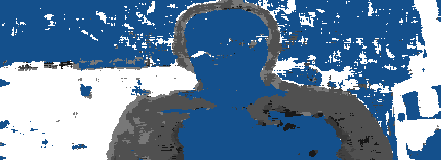}&
\includegraphics[width=86pt,height = 70pt]{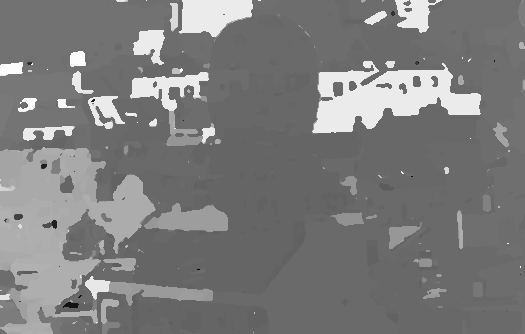}\\
(a)&(b)&(c)&(d)\\
\end{tabular}
\end{center}
\caption{Depth estimation: (a) Raw LF image. Depth map of (b) transmitted layer and (c) reflected layer (blue indicates presence of components from only one layer). (d) Estimated depth map using the technique of \cite{wang2015occlusion} \label{fig:real2}}
\end{figure*}
\section{Experimental results}

Firstly, we tested our layer separation algorithm on synthetic images while ignoring the effect of motion blur. For a set of depth values ranging from 0.35m to 2.3m (in steps of 0.15m), we simulated LF observations and reconstructed the texture volume. We used four different natural textures. We assumed that the true depth values were known. A representative example is shown in Fig. \ref{fig:corr}. To quantify the performance we evaluate the normalized cross correlation (NCC) between the true and estimated textures of both the layers and average it over all the four pairs of textures. Fig. \ref{fig:corr} (e) shows plots of mean NCC for cases wherein one of the layers had depth values fixed at $35$ cm, $80$ cm, and $1.7$ m. From the plot, it is clear that the performance of layer separation is good only when the depths of the two layers are far apart. We also note that at $50$ cm, there is a dip in the score for all three plots. This is because, this depth value was close to the camera main lens focal plane, wherein it is not possible to recover high resolution texture \cite{Bishop}. For synthetic experiments with motion blur, the average NCC was $0.897$ when evaluated on the result of joint motion deblurring and layer separation (see the supplementary material for further details).


\begin{figure*}[t]
\begin{center}
\begin{tabular}{cccc}
\includegraphics[width=86pt,height = 70pt]{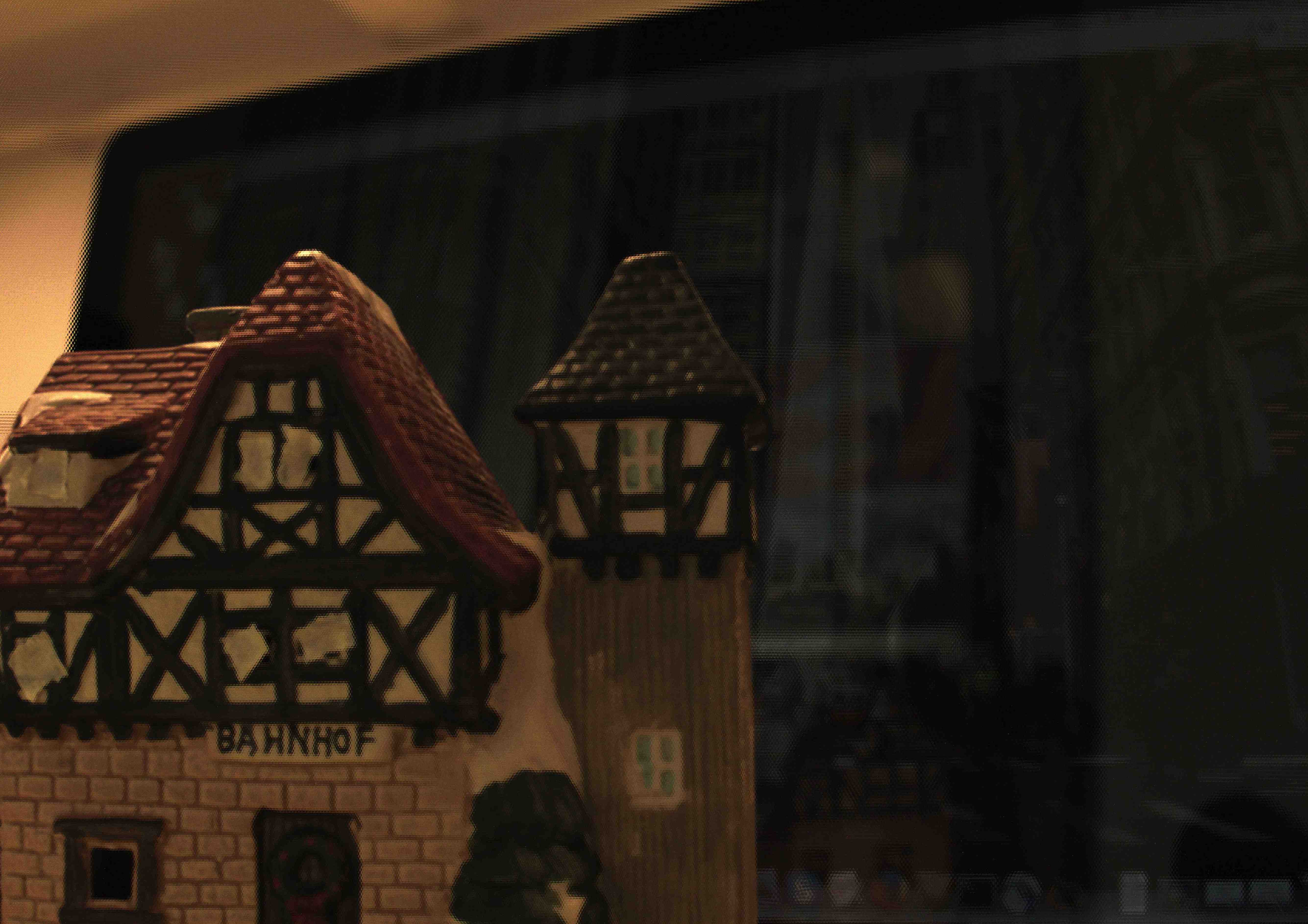} &
\includegraphics[width=86pt,height = 70pt]{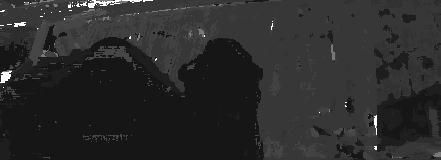}&
\includegraphics[width=86pt,height = 70pt]{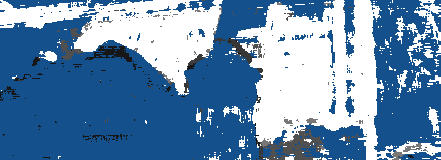}&
\includegraphics[width=86pt,height = 70pt]{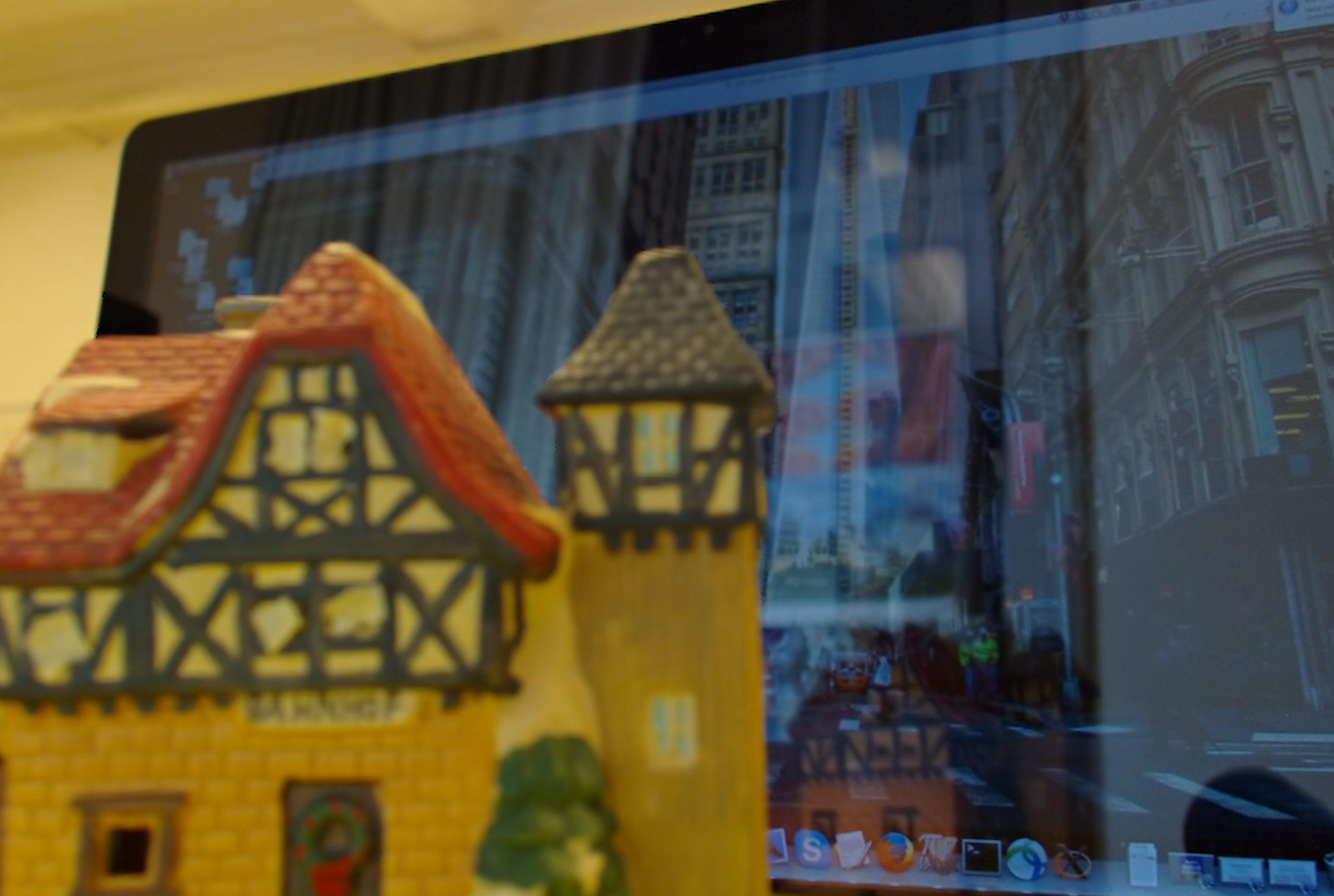}\\
(a)&(b)&(c)&(d)\\
\includegraphics[width=86pt,height = 70pt]{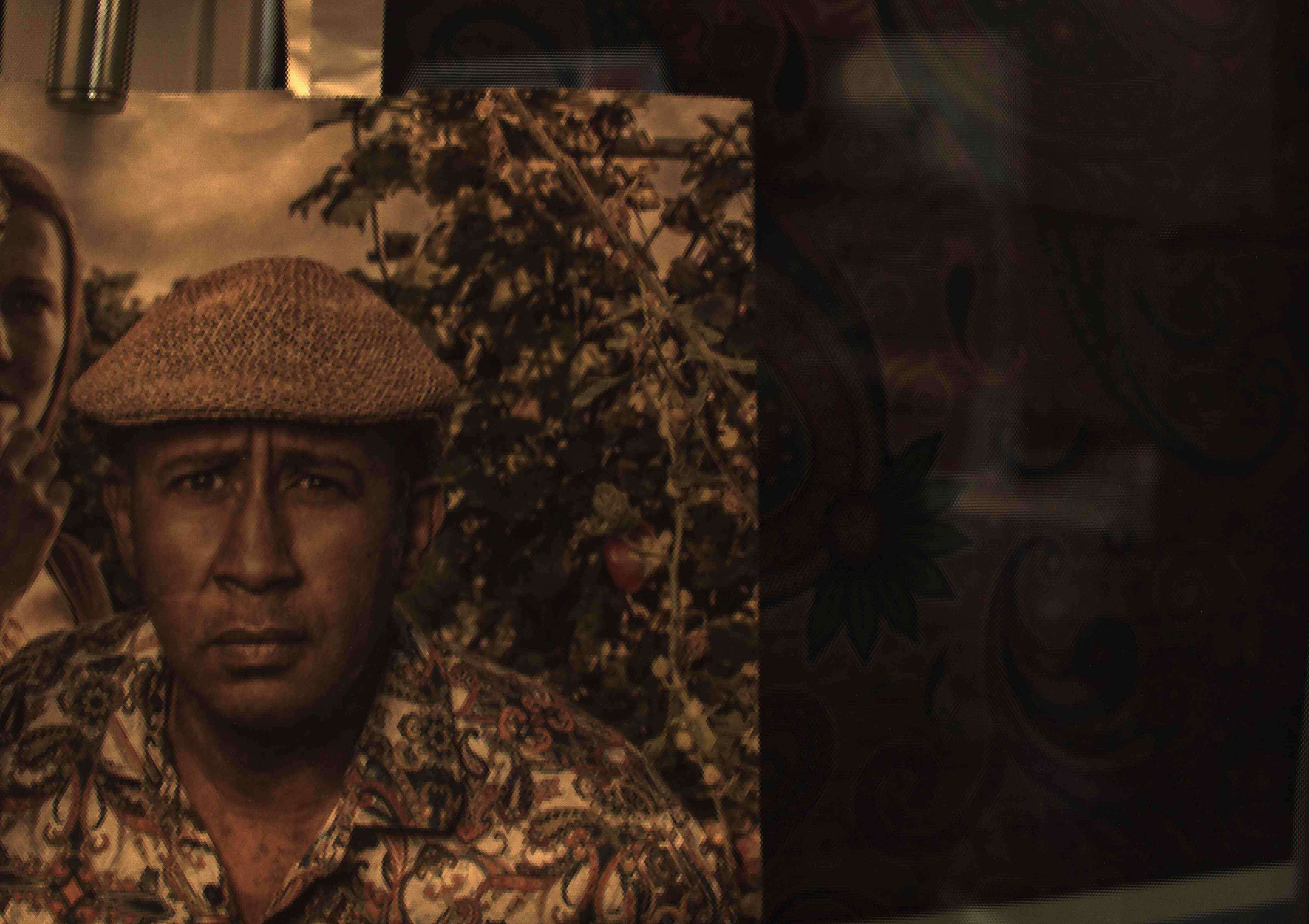} &
\includegraphics[width=86pt,height = 70pt]{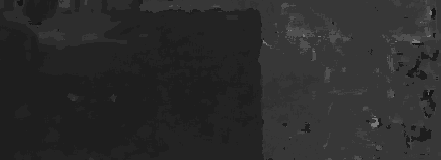}&
\includegraphics[width=86pt,height = 70pt]{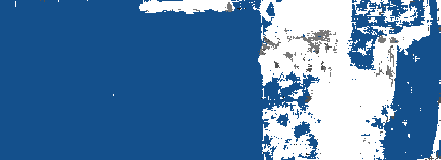}&
\includegraphics[width=86pt,height = 70pt]{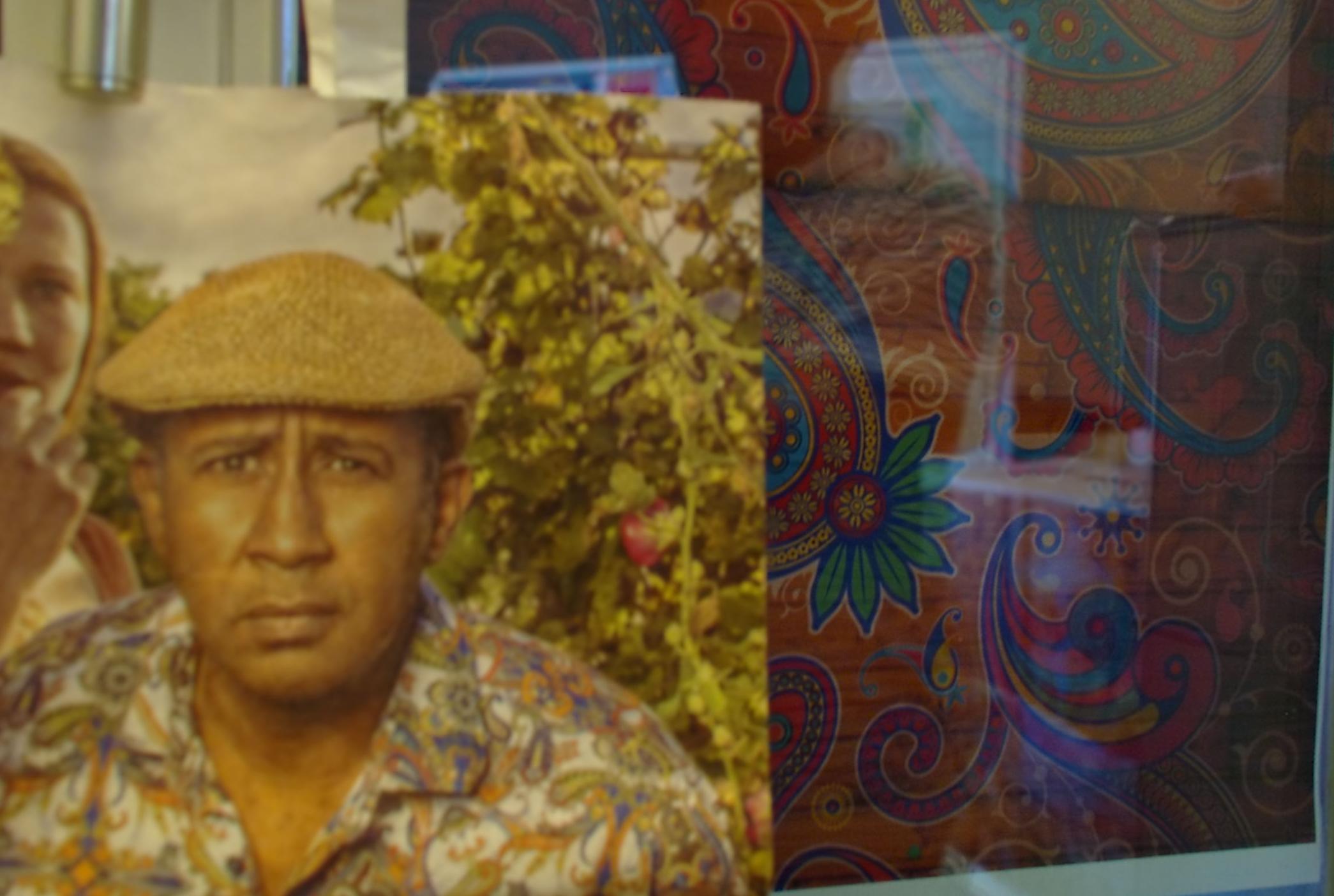}\\
(e)&(f)&(g)&(h)\\
\includegraphics[width=86pt,height = 70pt]{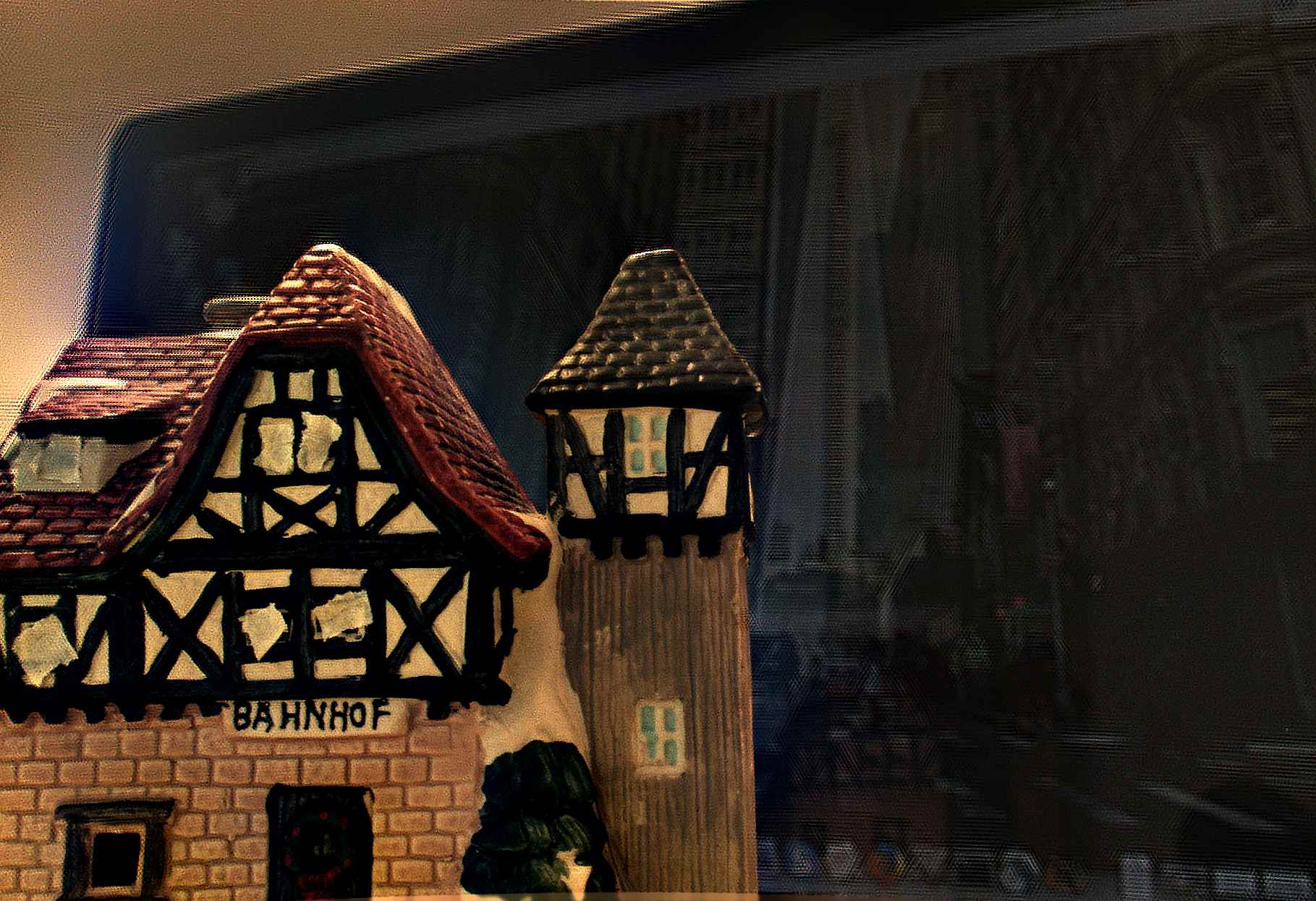} &
\includegraphics[width=86pt,height = 70pt]{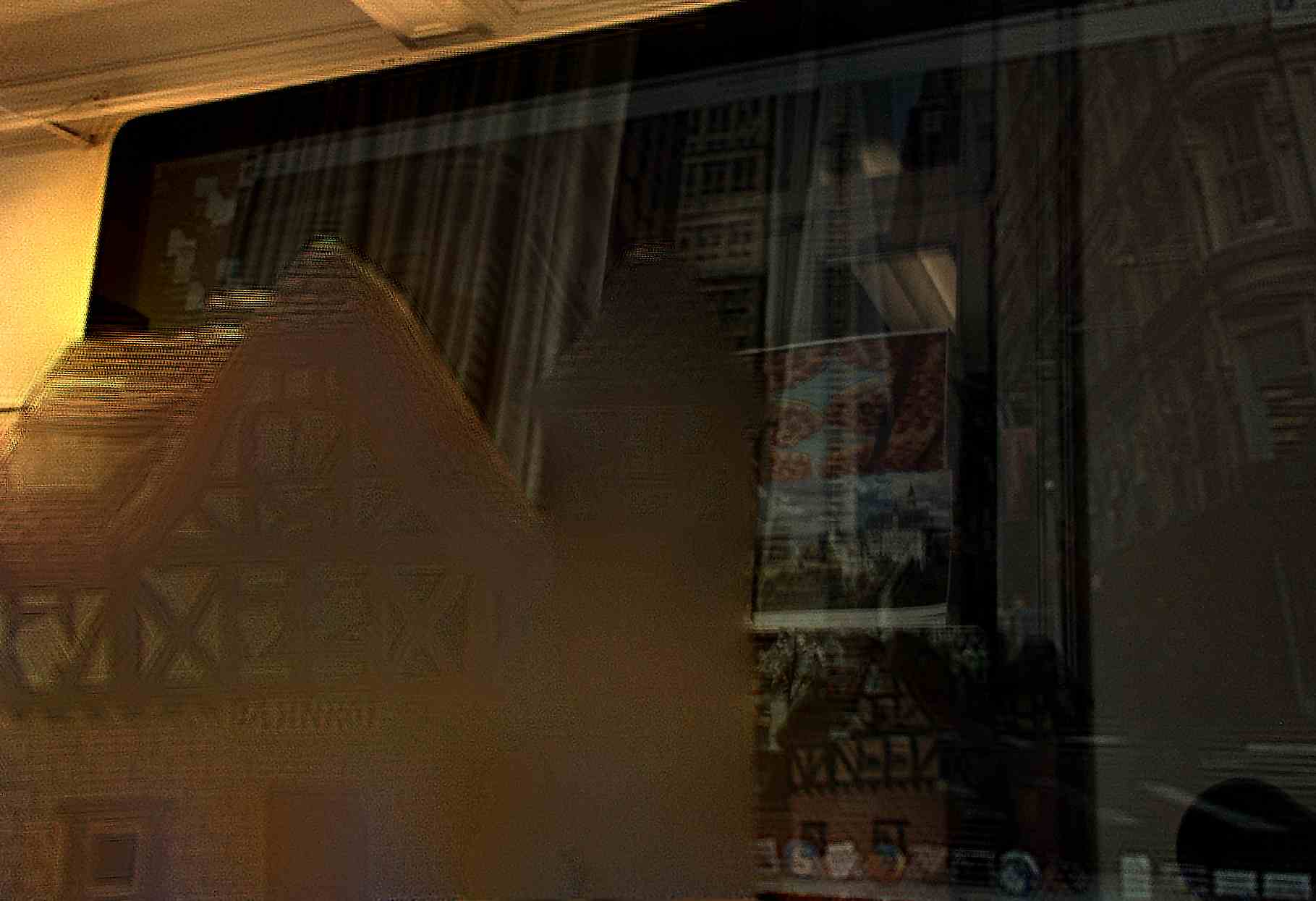}&
\includegraphics[width=86pt,height = 70pt]{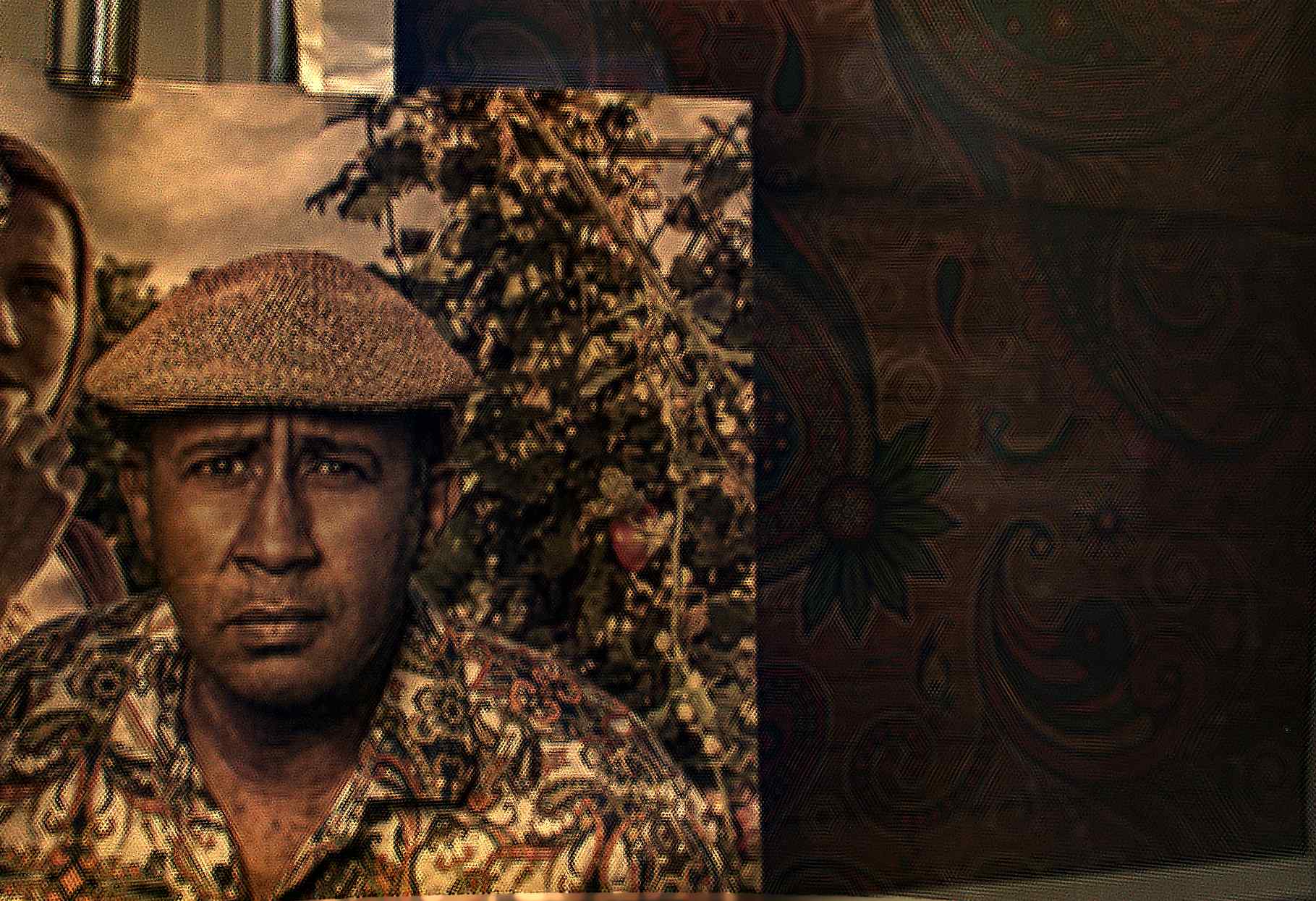}&
\includegraphics[width=86pt,height = 70pt]{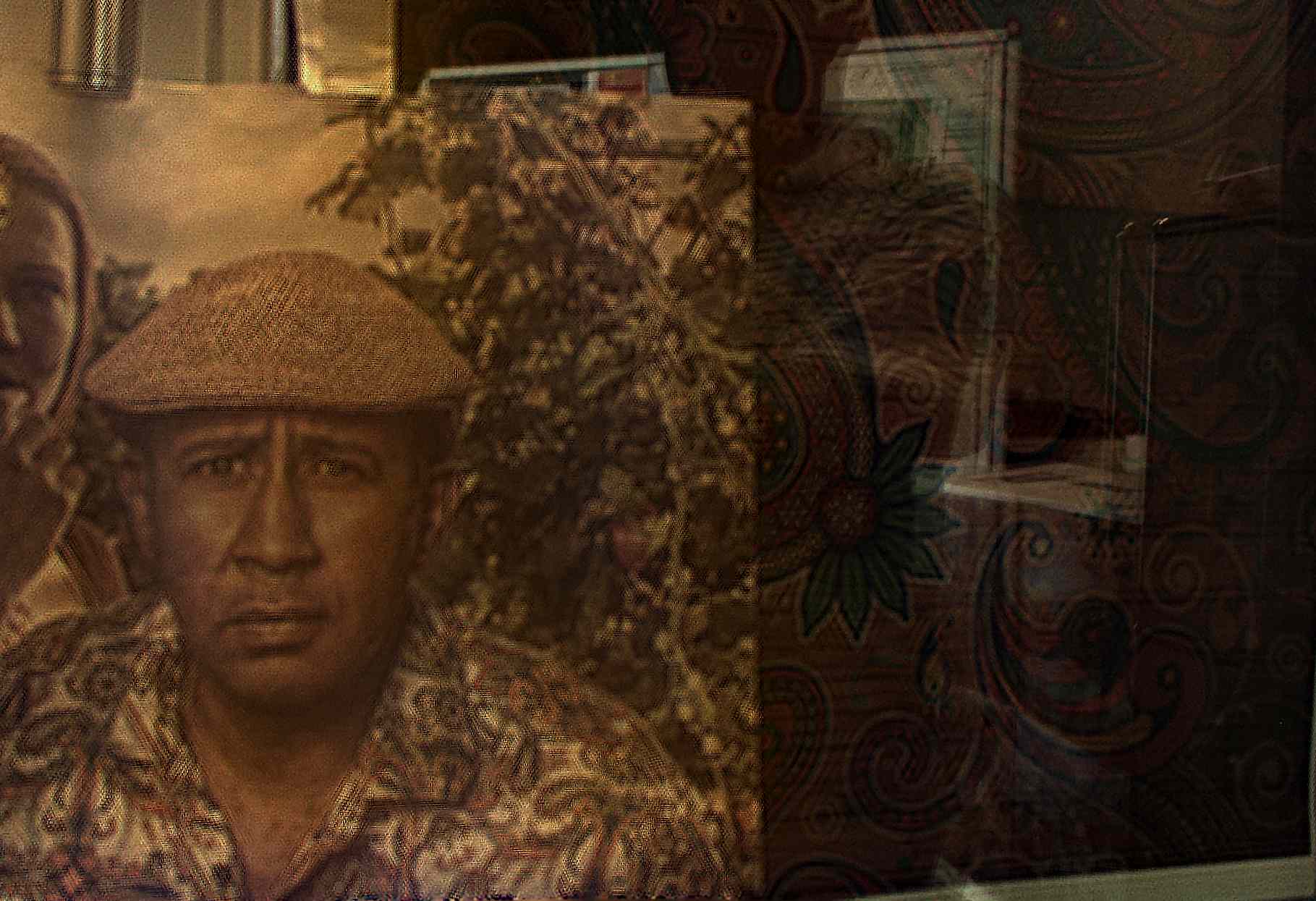}\\
(i)&(j)&(k)&(l)\\
\end{tabular}
\end{center}
\caption{Depth estimation: (a) and (e) Raw LF images. (b) and (f) Transmitted layer depth map (c) and (g) Reflected layer depth map (blue indicates presence of components from only one layer). (d) and (h) Lytro rendered image. (i)-(l) Recovered textures of the two layers. \label{fig:dep}}
\end{figure*}

We perform real experiments using the Lytro Illum camera. The scene in Fig.~\ref{fig:real1} consisted of a computer screen and a person holding the camera in a room. The estimated depth maps of the two layers are shown in Fig.~\ref{fig:real2} (b) and (c). The depth map of the transmitted layer is uniform throughout except for some artifacts, thereby correctly depicting the computer screen. In the second depth map, the blue color indicates the locations at which radiance components from only one layer were present. In the reflected layer, the regions corresponding to the white wall, person's face and the camera were completely textureless. Hence these locations were marked as regions without reflections. The depth map of the reflected layer correctly denotes the separation between the person and the background (refer to Fig.~\ref{fig:real1}). Fig.~\ref{fig:real2} (d) shows that depth estimation failed when reflections were not accounted for. Despite the assumption of constant depth for each layer, we see that the textures of the two layers have been well separated.

Fig. \ref{fig:dep} shows the result of depth estimation and layer separation on two other scenes that contained a Lambertian surface as well as a reflective surface. In both these images, our method correctly labels regions corresponding to Lambertian surfaces. In both these images, there are large regions with \emph{limited texture} in the reflected layer. Consequently, we see that even the textureless regions get marked as region without reflections. Note that 
in Fig. \ref{fig:dep} (c), the depth map correctly depicts the white regions corresponding to background which is far from the reflective surface and also shows a small grey region that corresponds to the reflection of the house model seen in the computer screen. In Figs. \ref{fig:dep}  (i)-(l), we show the recovered textures of the two layers corresponding to the two scenes. We observe artifacts in regions that are close to the camera. In this experiment, objects were as close as about $20$ cm from the camera. At this range, depth changes induce large disparity changes and therefore constant depth assumption for the entire layer can lead to artifacts.

\begin{figure*}[t]
\begin{center}
\begin{tabular}{ccc}
\includegraphics[width=117pt,height = 90pt]{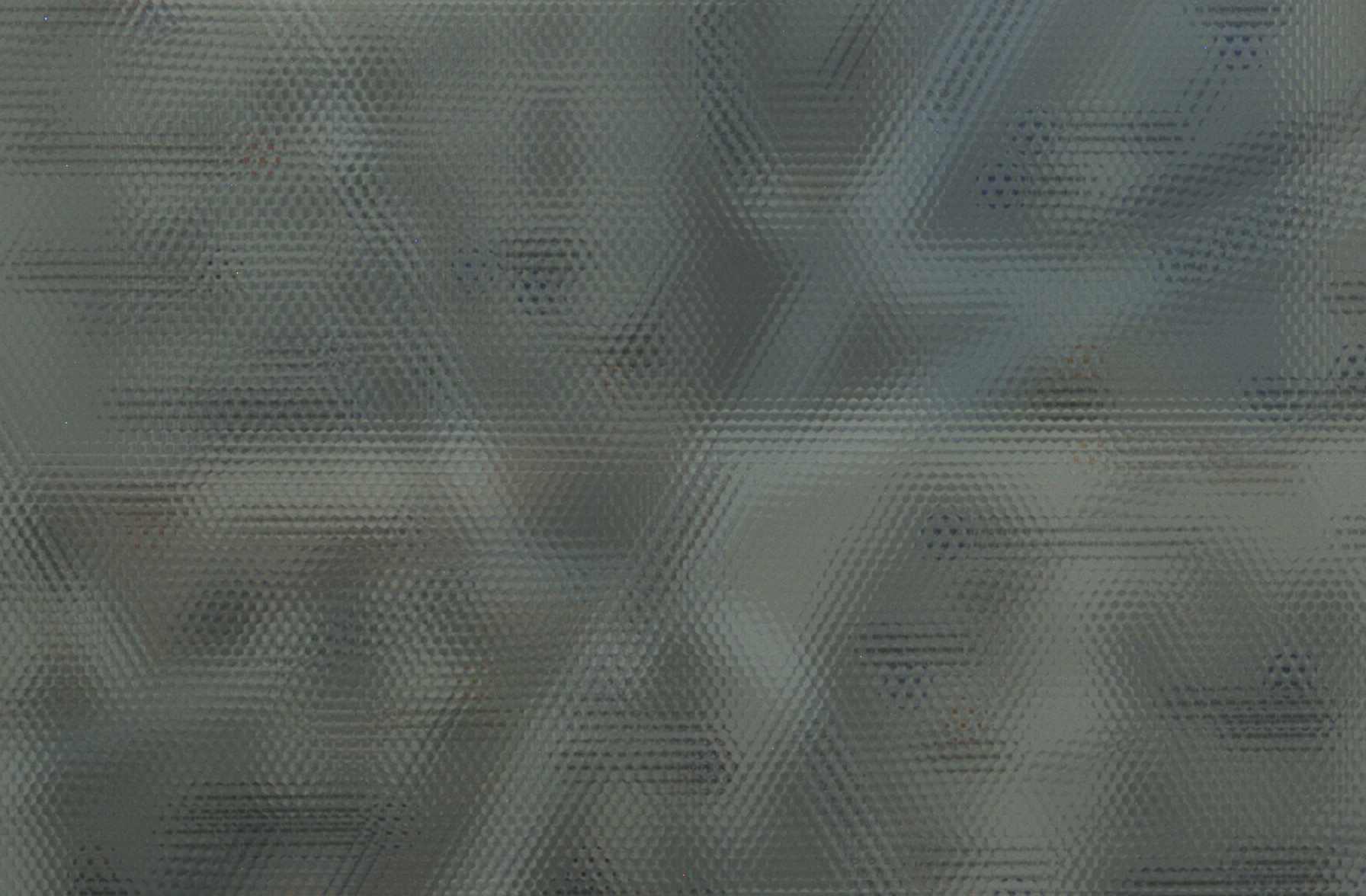}&
\includegraphics[width=117pt,height = 90pt]{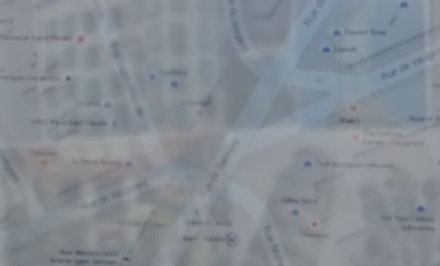}&
\includegraphics[width=117pt,height = 90pt]{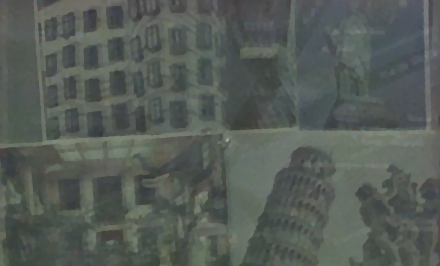} \\
(a)&(b)&(c)\\
\includegraphics[width=117pt,height = 90pt]{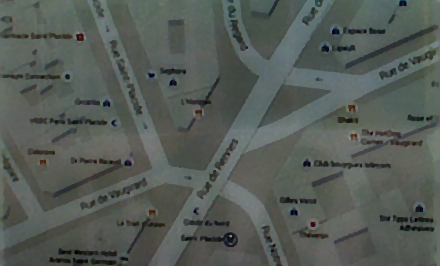}&
\includegraphics[width=117pt,height = 90pt]{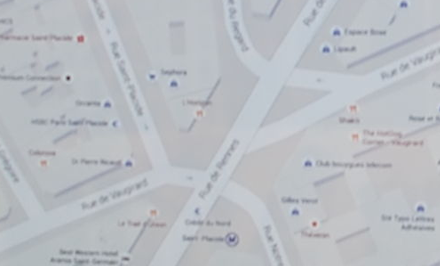} &
\includegraphics[width=117pt,height = 90pt]{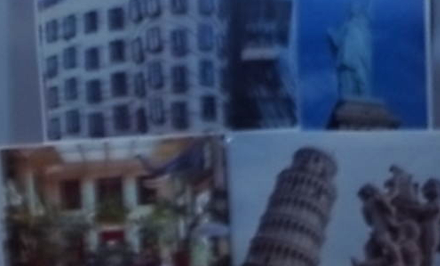}\\
(d)&(e)&(f)\\
\end{tabular}
\caption{(a) Raw image. (b) Lytro Desktop rendering of the observation. Recovered (c) reflected and (d) transmitted layer from the proposed scheme. Reference observations corresponding to (e) transmitted and (f) reflected layers \emph{without} superposition. \label{fig:room_ref1}}
\end{center}
\end{figure*}

\begin{figure*}[t]
\begin{center}
\begin{tabular}{ccc}
\includegraphics[width=117pt,height = 105pt]{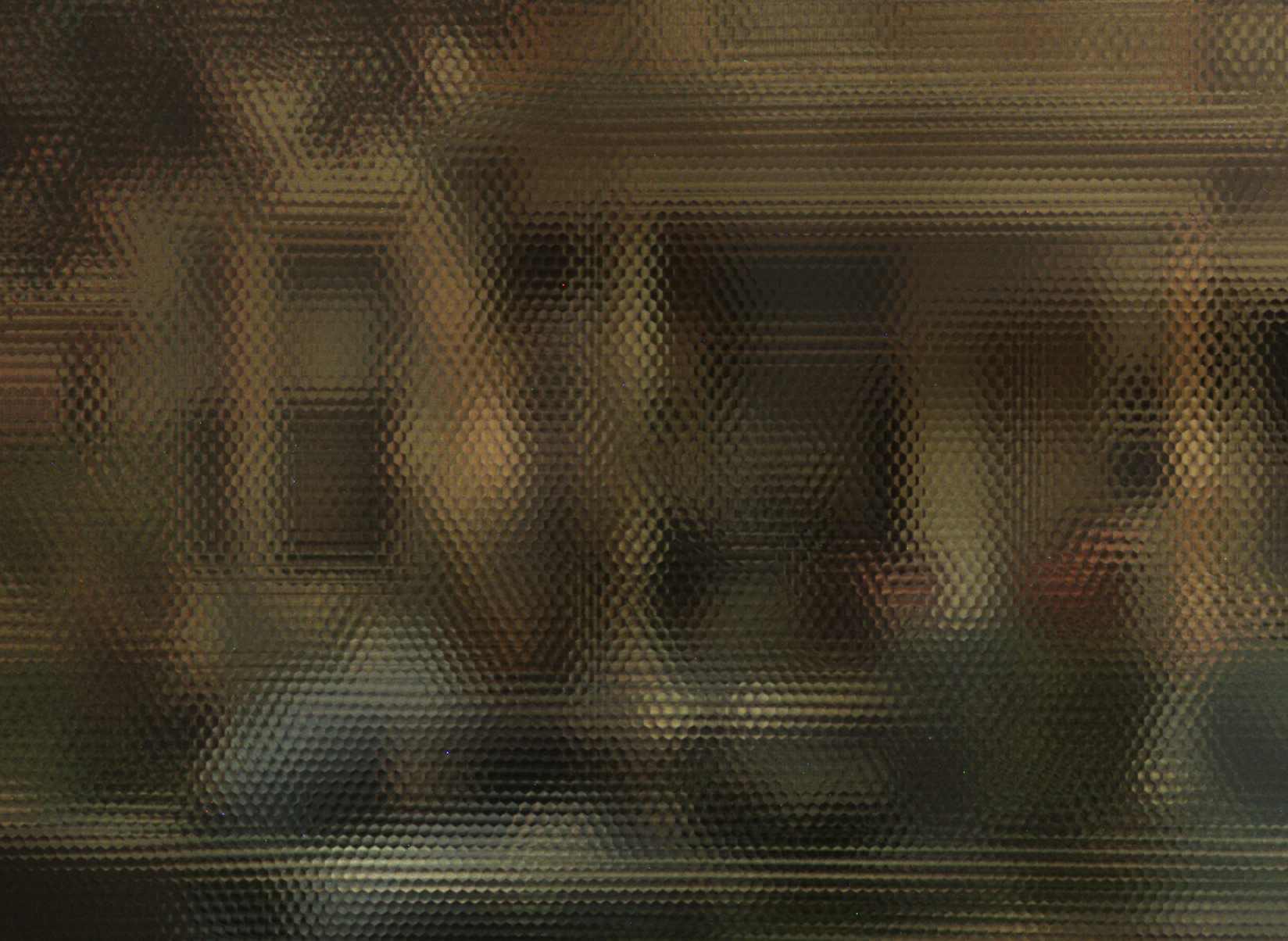} &
\includegraphics[width=117pt,height = 105pt]{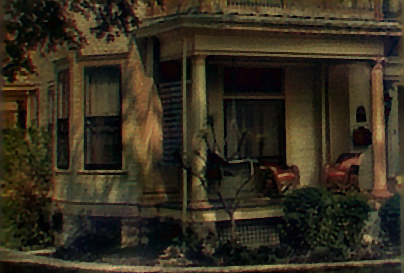}&
\includegraphics[width=117pt,height = 105pt]{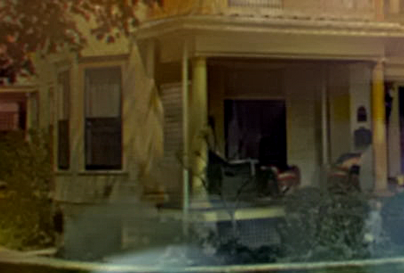}\\
(a)&(b)&(c)\\

\includegraphics[width=117pt,height = 105pt]{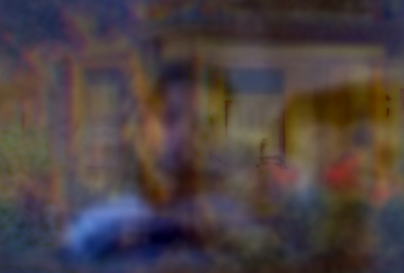}&
\includegraphics[width=117pt,height = 105pt]{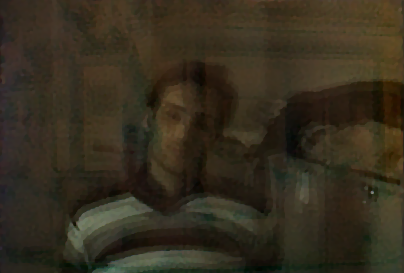}&
\includegraphics[width=117pt,height = 105pt]{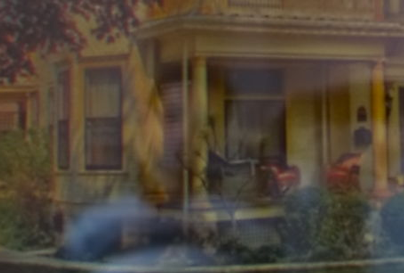} \\
(d)&(e)&(f)\\
\end{tabular}
\end{center}
\caption{Layer separation: (a) Raw LF image. (d) Rendered image by Lytro Desktop software. Estimated textures of (b) transmitted and (e) reflected layers from the proposed method. (c) and (f) show results of the algorithm in \cite{BrowCVPR14} on the Lytro rendered image.\label{fig:l1}}
\end{figure*}

In our next experiment, we illuminated a region in front of a computer screen and imaged the screen with a plenoptic camera. The rendering of the LF image from the Lytro Desktop software obtained by refocusing at the screen surface is shown in Fig. \ref{fig:room_ref1} (b). The result of the proposed depth estimation and layer separation method is shown in Figs. \ref{fig:room_ref1} (c) and (d). For purpose of comparison, we also captured images of the scene by avoiding the mixing of layers. The Lytro rendering of an image captured without illuminating the reflecting surface and by increasing the brightness of the computer screen is shown in Fig. \ref{fig:room_ref1} (e). The reference image for the reflected layer shown in Fig. \ref{fig:room_ref1} (f) was captured by turning off the screen. Note that in our result, the layers have been well separated even in the presence of textures with high-contrast. Furthermore, when one compares, Figs. \ref{fig:room_ref1} (d) and (e), our result (from the superimposed image) has a better resolution as against Lytro rendering (of the scene which did not have any layer superposition). Another real example is shown in Fig. \ref{fig:l1}. In Figs. \ref{fig:l1} (e) and (f) we see that the result obtained by applying the technique in \cite{BrowCVPR14} was not satisfactory.


\begin{figure*}
\begin{center}
\begin{tabular}{ccc}
\includegraphics[width=117pt,height = 105pt]{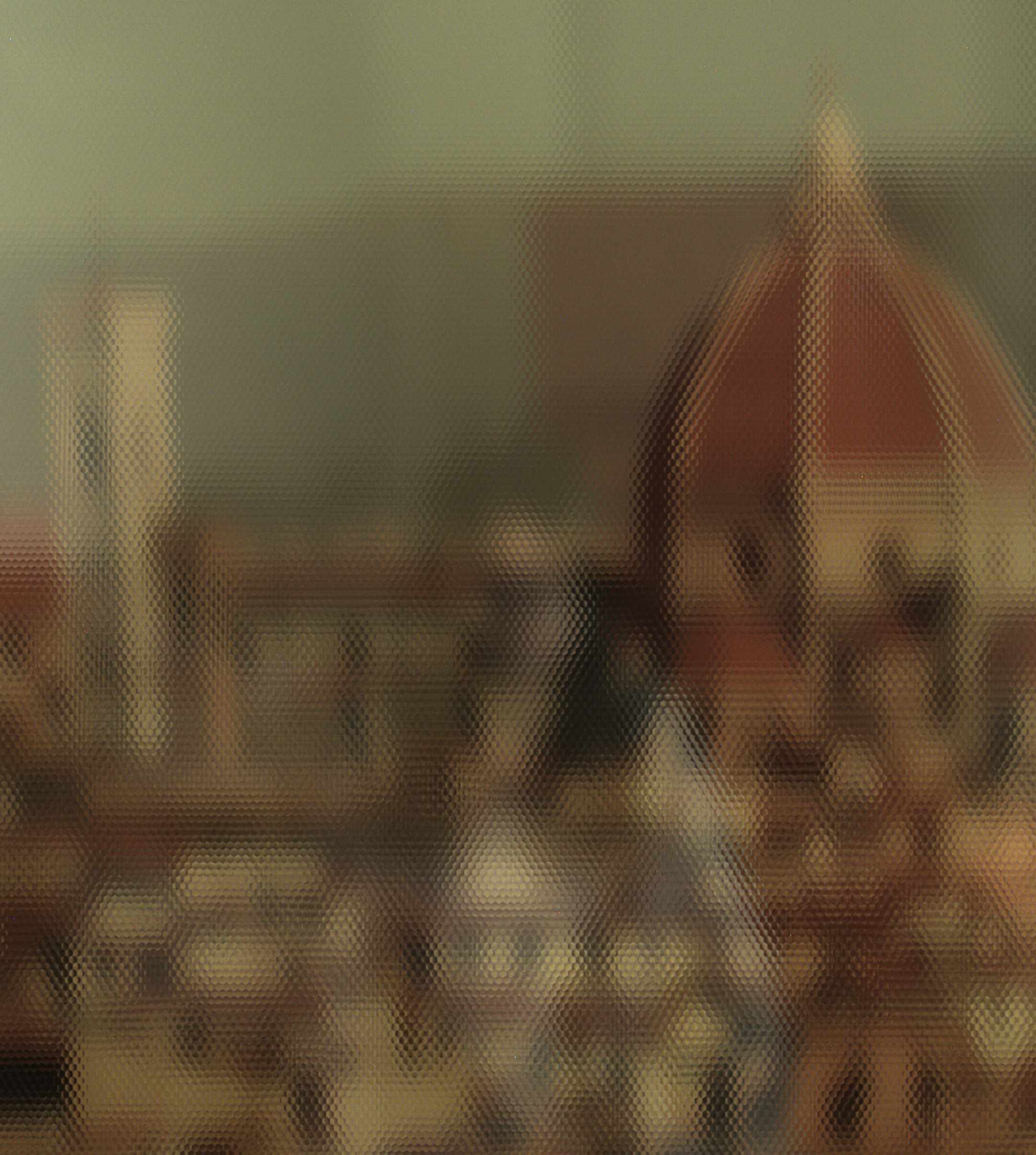}&
\includegraphics[width=117pt,height = 105pt]{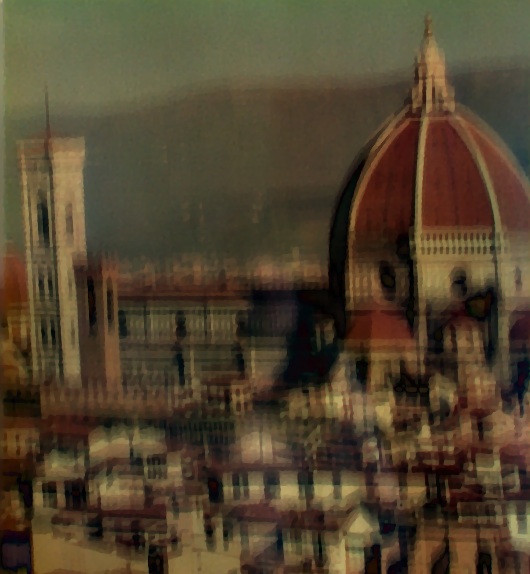}&
\includegraphics[width=117pt,height = 105pt]{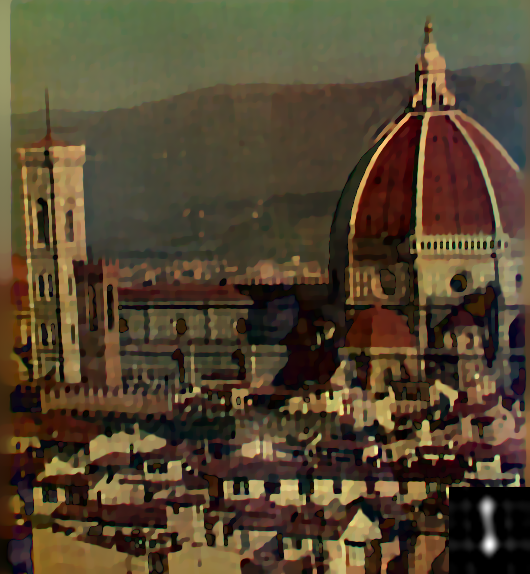} \\
(a)&(c)&(e)\\
\includegraphics[width=117pt,height = 105pt]{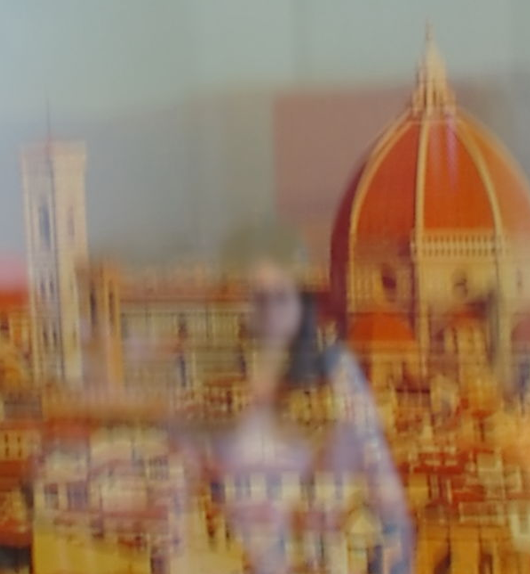}&
\includegraphics[width=117pt,height = 105pt]{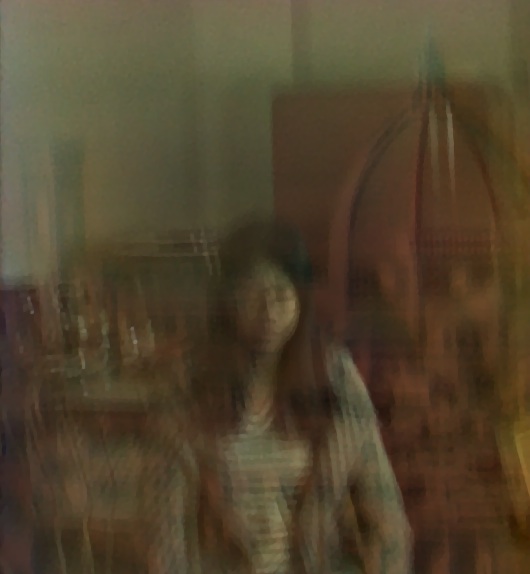} &
\includegraphics[width=117pt,height = 105pt]{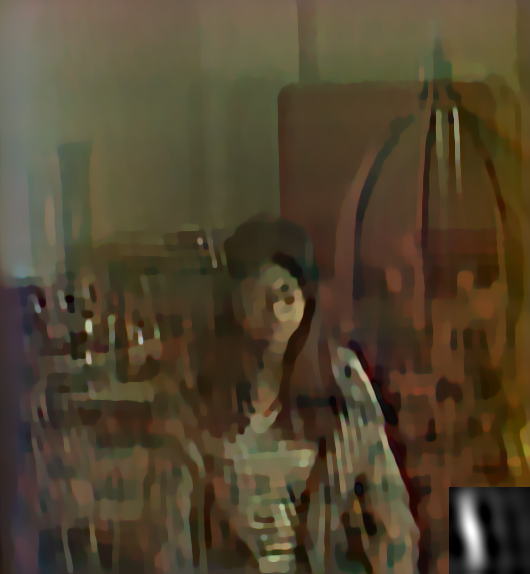}\\
(b)&(d)&(f)\\
\includegraphics[width=117pt,height = 105pt]{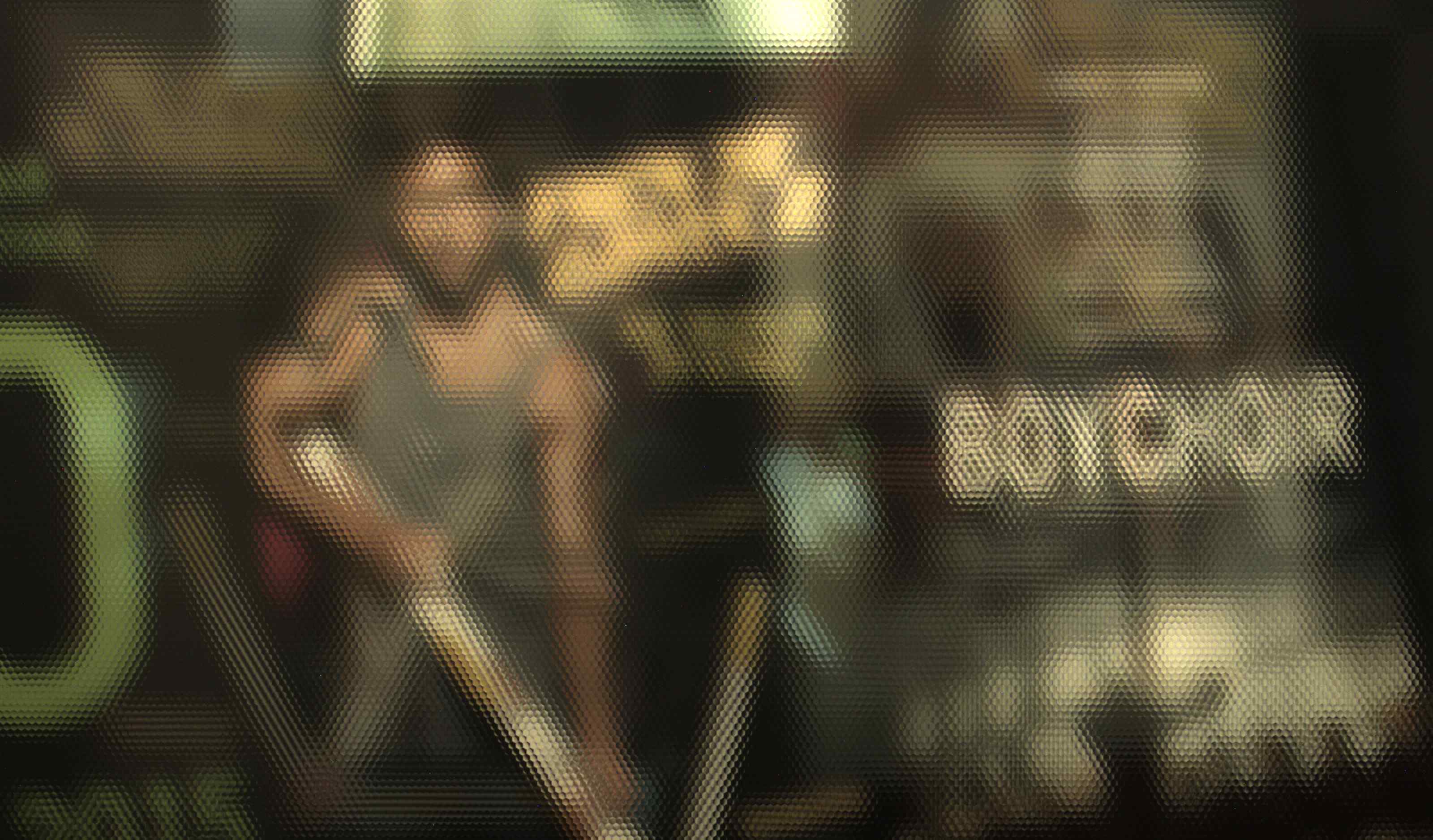}&
\includegraphics[width=117pt,height = 105pt]{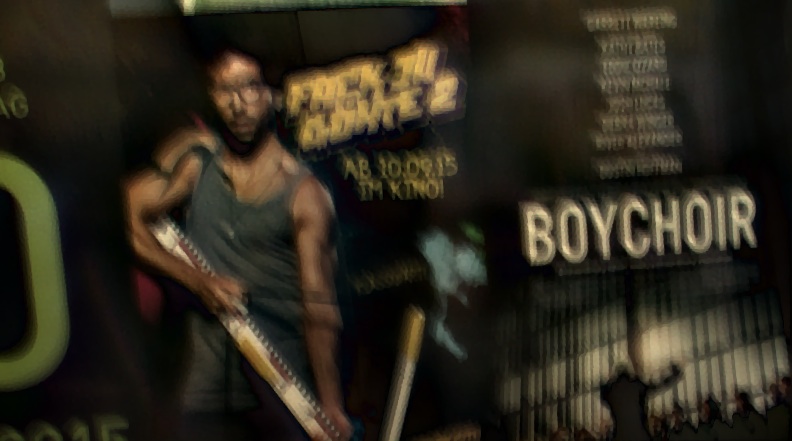}&
\includegraphics[width=117pt,height = 105pt]{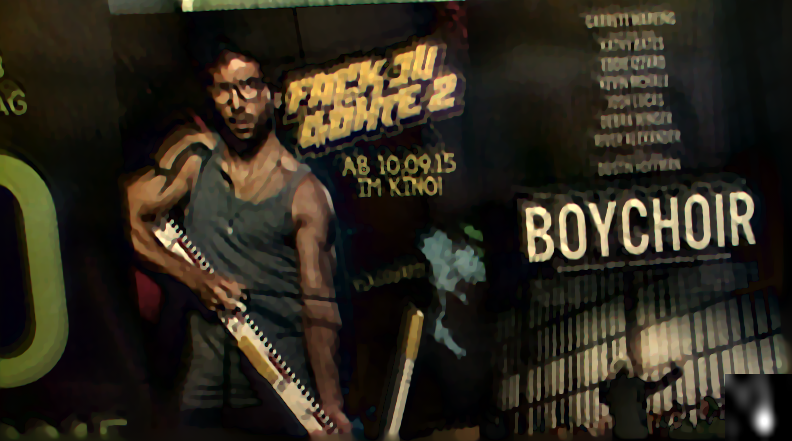} \\
(g)&(i)&(k)\\
\includegraphics[width=117pt,height = 105pt]{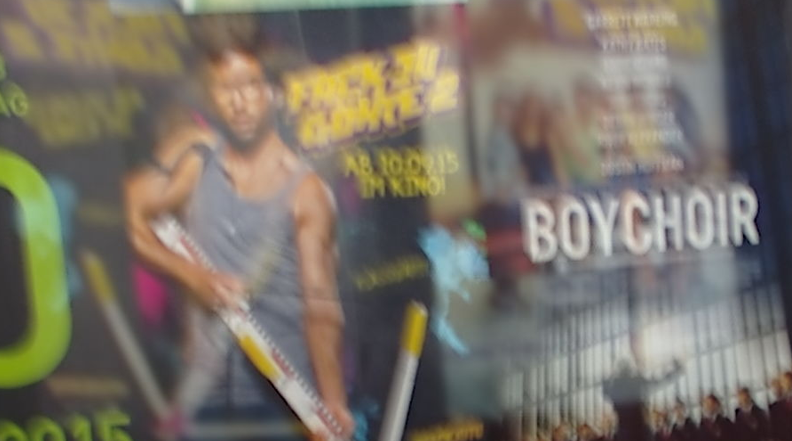}&
\includegraphics[width=117pt,height = 105pt]{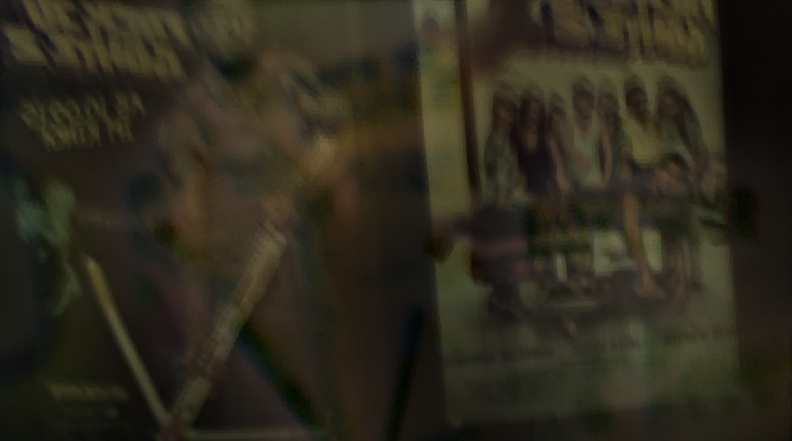} &
\includegraphics[width=117pt,height = 105pt]{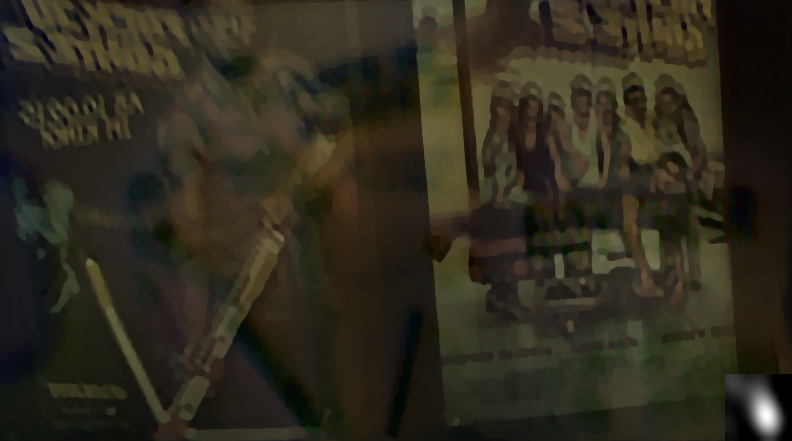}\\
(h)&(j)&(l)\\
\end{tabular}
\caption{Layer separation and motion deblurring: (a) and (g) Raw LF image. (b) and (h) Rendered image from Lytro software. Estimated textures of transmitted (c,i) and reflected (d,j) layers. Motion deblurred transmitted layer (e,k) and reflected layer (f,l) (with motion blur kernel shown as inset).\label{fig:kitch1}}
\end{center}
\end{figure*}

We next present results on motion blurred scenes captured with a handheld camera. In Fig. \ref{fig:kitch1}, we show the raw LF image, rendering by Lytro software (by manually refocusing at the transmitted layer), results of the proposed method with and without motion blur compensation. For the first scene, while the transmitted layer is that of a poster pasted at a window, the reflected layer is that of a person with different objects in the background. The second scene had movie posters in both the layers. In each of the two layers there was a poster with similar content. Through visual inspection of the results on these images we can see that our proposed method produces consistent results.

%
%
\section{Conclusions}
We developed a technique to address the depth estimation and layer separation problem from a single plenoptic image. Not only our ConvNet-based approach enables depth estimation, but also detects the presence of reflections. With the estimated depth values we demonstrated that our PSF-based model enforces constraints that render layer separation feasible. Within our framework we also addressed the challenging problem of motion deblurring. In texture reconstruction we considered that both layers have a constant depth. In the future, our objective is to relax this assumption. Moreover, the performance of the ConvNet-based classifier can be further improved by having more variations in the training data and by including more depth labels.

\bibliographystyle{splncs}
\bibliography{reflection,papers,egbiblf}


\end{document}